\documentclass[a4paper]{cas-sc}

\def\tsc#1{\csdef{#1}{\textsc{\lowercase{#1}}\xspace}}
\tsc{WGM}
\tsc{QE}

\usepackage{graphicx}%
\usepackage{multirow}%
\usepackage{amsmath,amssymb,amsfonts}%
\usepackage{amsthm}%
\usepackage{mathrsfs}%
\usepackage[title]{appendix}%
\usepackage{xcolor}%
\def\BibTeX{{\rm B\kern-.05em{\sc i\kern-.025em b}\kern-.08em
		T\kern-.1667em\lower.7ex\hbox{E}\kern-.125emX}}

\usepackage{textcomp}%
\usepackage{manyfoot}%
\usepackage{booktabs}%
\usepackage{algorithm}%
\usepackage{algorithmicx}%
\usepackage{algpseudocode}%
\usepackage{listings}%

\usepackage[authoryear,longnamesfirst]{natbib}
\usepackage{hyperref}

\usepackage{textcomp}
\usepackage{diagbox} 
\usepackage{siunitx} 
\usepackage{caption} 
\usepackage{adjustbox}

\theoremstyle{thmstyleone}%
\theoremstyle{thmstyletwo}%

\theoremstyle{thmstylethree}%

\begin{document}
\let\WriteBookmarks\relax
\def\floatpagepagefraction{1}
\def\textpagefraction{.001}

\shorttitle{ForestNN}    

\shortauthors{A.~Nazari~et~al.}  

\title [mode = title]{Kinship Verification through a Forest Neural Network}  



\author{Ali Nazari}[orcid=0000-0003-3938-1480]
\ead{al_nazari@sbu.ac.ir}
\credit{Conceptualization, Data Curation, Formal Analysis, Investigation, Methodology, Software, Validation, Visualization, Writing - Original Draft, Writing - Review \& Editing}

\author{Omidreza Borzoei}[orcid=0009-0006-9779-205X]
\ead{borzoei.omidreza@gmail.com}
\credit{Experiments, Parts of Software, Writing - Review \& Editing}

\author{Mohsen Ebrahimi Moghaddam}[orcid=0000-0002-7391-508X]
\ead{m_moghadam@sbu.ac.ir}
\credit{Supervision, Project Administration, Writing – review \& editing}
\cormark[1]
\cortext[cor1]{Corresponding author}

\affiliation{organization={Faculty of Computer Science and Engineering, Shahid Beheshti University},
	addressline={Velenjak}, 
	city={Tehran},
	postcode={1983969411}, 
	state={Tehran},
	country={Iran}}


\begin{abstract}
Early methods of kinship verification relied on face representations, which were less accurate than joint representations of parents' and children's facial images learned from scratch. 
We propose a new approach based on graph neural network concepts to leverage face representations, including a specific classification module that enables our network to be trained not only with common losses in kinship verification but also with the center loss, gradually introduced as accuracy increases after the early epochs.
Furthermore, we achieved higher results on the KinFaceW-II dataset, with average improvements of nearly 1.67 and 0.8 across all kinship types compared with D4ML and DSMM, respectively. Specifically, there were statistically significant improvements on the F-S ($p<0.01$), F-D ($p<0.01$), and M-S ($p<0.05$) relations on KinFaceW-II compared with D4ML. We were also close to the best performance of the cutting-edge algorithms on the KinFaceW-I dataset. 
Those results ultimately demonstrate the effectiveness of our approach and achieve accuracy comparable to that of joint representation algorithms.
\end{abstract}


\begin{highlights}
\item We propose the Forest Neural Network (FNN), a GNN, to model kinship relations. It considers the disjoint nature of face components and exchanges information among them via GNN operators to bridge the gap between facial and kinship representations.

\item We train our FNN model using facial features to reduce computation and leverage multiple CNNs, each serving as an expert system to improve KV accuracy. We also redesigned an attention-based CNN, the Extended Residual Network (ERN), to leverage features of high-resolution images produced from low-resolution ones.

\item A new concerted combination of losses is introduced so that the center loss, along with other KV losses, can condense the features of each class, facilitating easier classification and outperforming previous studies. This fusion tries to overcome the slight inter-class difference and intra-class variance.
\end{highlights}


\maketitle

\section{Introduction}
Visual Kinship Verification (KV) takes two face images to determine whether they share a familial relationship. It has considerable potential usage in sociology, anthropology, psychology, neuroscience, genetics, privacy protection, and social media~\cite{mzoughi2024review,wang2023survey,wu2022facial,robinson2021survey} 
for example, finding missing family members~\cite{xia2012understanding}, collecting photos for the family album~\cite{robinson2021survey}, family privacy protection~\cite{kumar2020adversary}, border control and criminal investigation~\cite{mzoughi2024review,lu2013neighborhood}. Moreover, it has been recently investigated in smart environments to recognize family members~\cite{bisogni2022kinship}.

Furthermore, KV challenges can be categorized into two parts: first, some are common to other computer vision algorithms, such as the small size of KV datasets, pose and lighting variations, diverse sources, facial expression variations, occlusions, blurring, and low-resolution images. Additionally, some challenges are unique to KV, such as age variation, similarity between family members and strangers (slight inter-class difference), environmental effects on family members' faces, dissimilarity among some family members (intra-class variance), limited training data~\cite{mzoughi2024review,wang2023survey,wu2022facial,robinson2021survey}, and multiple feature fusion~\cite{liang2018weighted}.

Generally, two approaches tackle the problem of kinship verification in the presence of all challenges: 1. face matching (joint representation) and 2. face representation-based (recognition-based) KV methods.
In the face matching or joint representation, two face images are processed jointly to determine their kinship relation~\cite{li2021reasoning}. The most recent research has investigated this approach and achieved the most advanced and competitive results in the field, often referred to as state-of-the-art~\cite {li2021reasoning,zhu2023distance}. 

On the other hand, most early methods fell into the second approach and first analyzed faces to obtain unique features as face representations. Then, their algorithms decided their kinship relation based on those face representations. One of the substantial benefits of this approach is that the KV algorithm can be used simultaneously with highly demanding applications, specifically face recognition (FR). The other advantage of this approach is that it does not require retraining a model from facial images, since we already have a representation of each face. The drawback of face representation-based KV algorithms is that they are not as accurate as face matching methods~\cite{kohli2016hierarchical}.

Ultimately, our purpose is to propose a face representation-based KV method with high accuracy, leverage Graph Neural Network (GNN) concepts to model family relations, and apply facial features to train our model, the Forest Neural Network (FNN).

To our knowledge, this paper is the first to apply graph neural network concepts to kinship verification. There are rare papers that use graph structures in kinship verification. Graph embedding-based metric learning~\cite{liang2018weighted} introduced an intrinsic graph and two penalty graphs to acquire multiple metrics. H-RGN~\cite{li2021reasoning} designed a tree-like structure by grouping feature elements as nodes and performing convolutional operations among them. Although these papers attempted to apply graph concepts, they did not completely use GNN concepts.

Our key contributions to elevate the accuracy of kinship verification are enumerated as follows:
\begin{itemize}
	\item We propose the Forest Neural Network (FNN), a GNN, to model kinship relations. It considers the disjoint nature of face components and exchanges information among them via GNN operators to bridge the gap between facial and kinship representations.
	
	\item We train our FNN model using facial features to reduce computation and leverage multiple CNNs, each serving as an expert system to improve KV accuracy. We also redesigned an attention-based CNN, the Extended Residual Network (ERN), to leverage features of high-resolution images produced from low-resolution ones.
	
	\item A new concerted combination of losses is introduced so that the center loss, along with other KV losses, can condense the features of each class, facilitating easier classification and outperforming previous studies. This fusion tries to overcome the slight inter-class difference and intra-class variance.
\end{itemize}

Eventually, the organization of this paper is as follows:
Section~\ref{sec:related-work} briefly expresses related work. The proposed algorithm is explained in Section~\ref{sec:proposed-method}. Section~\ref{sec:experiment} provides experiments, their configuration, and the analysis sub-section to clarify various parts of the proposed algorithm. Eventually, Section~\ref{sec:conclusion} concludes the paper.

\section{Related Work}
\label{sec:related-work}
Researchers have proposed various approaches to address kinship verification, ranging from low- and mid-level features to algorithms such as traditional machine learning, metric learning, deep learning, and hybrid models~\cite{mzoughi2024review,wu2022facial}. 

Additionally, they used detailed, complete facial information to facilitate face analysis in early methods. Hence, they attempted to discover a description of faces, especially discriminatory features, using low-level features such as facial distances, gradient histograms, and color information to determine kinship relations~\cite {fang2010towards}. 

Furthermore, textural features, LBP variants~\cite{bottinok2015multi}, Gabor filter~\cite{zhou2012gabor}, salient facial features~\cite{guo2012kinship}, self-similarity~\cite{kohli2012self}, and SIFT Flow~\cite{puthenputhussery2016sift} were explored. Most of them used SVM as a classifier in the final stage.
In a study \cite{wang2015leveraging}, appearance and geometric features were used to separate family photos from non-family photos. Facial landmarks were employed to compute distances between two face images~\cite{wang2015leveraging}. From these distances, two groups of code-book associated with different age gaps were generated to distinguish family photos.
In recent work~\cite{nader2024enhanced}, color and texture handcrafted features such as Heterogeneous Auto-similarities of Characteristics (HASC), Color Correlogram (CC), and Dense Color Histogram (DCH) were combined to improve accuracy.

Regarding mid-level features, they are between high-level~(labels or values significant in making decisions) and low-level ones; they are more robust because low-level features are directly related to pixel values and not robust to variations. Mid-level features are constructed from low-level ones such as SPLE~\cite{zhou2011kinship} and Canonical correlation analysis~\cite{chen2017kinship}. Despite the mid-level methods to obtain values, attributed-based methods attempted to acquire description features such as narrow eyes and oval faces~\cite {xia2012toward, zhang2014lift}.

Another KV track explores metric learning algorithms that compute greater similarity between kin-related faces than between non-kin faces. One of the first activities in this approach was~\cite{somanath2012can} to learn an ensemble of kernels for each class. NRML~\cite{lu2013neighborhood} attempted to make closer positive samples(kin faces) and repel negative samples(non-kin faces). It iteratively computed this action and weighed the different importance of various negative samples. Ensemble similarity learning~\cite{zhou2016ensemble} put forward a sparse similarity function to consider attributes in kin samples. Some efforts~\cite{hu2017sharable} learned multiple metrics from different feature views.

As deep learning has gained superiority in discovering higher-level features and achieving greater accuracy, the most recent research has focused on this approach. In~\cite{dehghan2014look}, paired face images were patchified; the patches were fed into an auto-encoder to obtain features, which were then mapped to a space and ultimately to a classifier. An auto-encoder~\cite {kohli2018supervised} was also used to obtain class-specific representative features.

The paper~\cite{zhang122015kinship} split face images into 10 patches using their landmarks. These patches were fed into a basic CNN, comprising three convolutional layers, and their outputs were sent to a classifier. Their mean accuracies on KinFaceW-I and KinFaceW-II were 77.5$\%$ and 88.4$\%$, respectively, across all relationships.

In addition, a large kinship dataset, known as Families In the Wild (FIW)~\cite{robinson2016families}, consisting of 1000 families, was introduced. They fine-tuned two VGG networks, one with a triplet loss and another with a softmax loss, and reduced the feature dimensions of two fine-tuned networks using PCA, and achieved an average accuracy of 71.2$\%$ on the FIW dataset.

The adaptation approach (transfer learning) from a face dataset to a kinship dataset was proposed in~\cite{duan2017face}. Its CNN consisted of five convolution layers and one fully connected (FC) layer. It fine-tuned paired face images on their first CNN network, a coarse CNN with 1500 neurons in the last FC layer, and then they built two CNN networks from this coarse CNN network. These two CNN networks were again trained on Kin pairs. Their difference was that in one, all layers were trained entirely, whereas in the other, only the FC layer was trained. Eventually, NRML was applied to the features of these two networks.

Regarding recent methods, an attention-based architecture was proposed in~\cite{yan2019learning} to extract local information from face components. OR$^2$Net~\cite{li2024or2net} employed hard negative samples beyond the predefined configuration protocol to leverage information from all negative samples by reweighting them. 
FaCoRNet~\cite{su2023kinship} utilized the contrastive loss and cross-attention mechanism to pull closer facial components of kin images and pull away components of non-kin images.
The research~\cite{oruganti2024kinship} introduced the use of the curvelet transform (CLT) followed by a CNN for kinship verification of children's images.
The paper~\cite{nader2025efficient}, which we called FUSE, combined features of the ConvNext, EfficientNet, and ViT models to achieve higher results. GADA~\cite{zhao2026generative} generated facial images at various ages by a diffusion-based model. Then, the facial regions of these synthesized images were blended using sinc-based Lanczos interpolation and Poisson blending. The consistency loss was employed to reduce the gap between the original and synthesized child image and the parent or grandparent image. 

Some research has been conducted on combining the aforementioned approaches, known as Hybrid Modeling~\cite{mzoughi2024review}, which can be an ensemble of the same models or a combination of diverse models. Ensemble similarity learning (ESL)~\cite{zhou2016ensemble} learned a similarity function and then used it in different learners. Finally, the results of these learners were merged. JLNnet~\cite{wang2020kinship} utilized an ensemble of KVs for kinship identification. A summary of the strengths and drawbacks of kinship verification methods is provided in Table~\ref{table:summary-methods}.

\begin{table}[htbp]
	\caption{A summary of the strengths and drawbacks of kinship verification methods}
	\label{table:summary-methods}
	\begin{tabular}{lll}
		methods            & strengths  & drawbacks \\ \hline
		
		Low-level features 
		& \begin{tabular}[c]{@{}l@{}}
			Simple and interpretable, \\ 
			encoding local information, \\ 
			easy to implement
		\end{tabular}                                         
		& \begin{tabular}[c]{@{}l@{}}
			Hand-crafted features, \\ 
			Sensitive to small variations, \\ 
			a lack of familial relation\\
			understanding, no robustness\\
			to real-world images,~and \\
			limited generalization
		\end{tabular}
		\\ \hline
		
		mid-level features 
		& \begin{tabular}[c]{@{}l@{}}
			More robust to variations,\\
			more familial relation\\
			understanding than low-level\\ features 
		\end{tabular}
		& \begin{tabular}[c]{@{}l@{}}
			Sensitive to environmental\\ 
			factors such as facial \\
			accessories, age gap, lighting
		\end{tabular}
		\\ \hline
		
		Metric learning
		& \begin{tabular}[c]{@{}l@{}}
			Learn similarity space, \\
			learn discriminative and \\
			distance-based features 
		\end{tabular}
		& \begin{tabular}[c]{@{}l@{}}
			Imbalanced samples (the number\\
			of non-kin pairs is huge),\\
			age gap, occlusion, accessories
		\end{tabular}
		\\ \hline
		
		Deep learning
		& \begin{tabular}[c]{@{}l@{}}
			Complex, nonlinear, high-level\\
			features, robust to visual variations,\\
			familial relation understanding,\\
			end-to-end learning, scalability,\\
			highest accuracy 
		\end{tabular}
		& \begin{tabular}[c]{@{}l@{}}
			Lack of large labeled datasets,\\
			dependency on augmentation, \\
			high computational cost, \\
			low interpretability, and\\
			lack of transparency
		\end{tabular}
		
	\end{tabular}
\end{table}

Furthermore, researchers have widely used multimodal data across various fields of computer vision. In KV, two multimodal datasets have been introduced, one that uses voice alongside images~\cite{wu2022audio} and another that contains video, audio, and text captions~\cite{robinson2021families}. 

In conclusion, we observe that cutting-edge algorithms~\cite{mzoughi2024review,wu2022facial,li2021reasoning,zhu2023distance} have recently been trained jointly on parent and child face images. These networks have learned a joint representation of parent-child images. 
Their results specify that early face representation-based methods were less accurate than their proposed method. Therefore, we propose a face-representation-based method that improves KV accuracy and applies graph neural network concepts to model family relations.

\section{Proposed Method}
\label{sec:proposed-method}

Our initial impression comes from some experiments in cognitive science that human beings look for facial features and signals to identify similarities among family members~\cite{hansen2020kin,cole2017human,dal2006kin}. Those findings show the importance of facial components in KV.
Additionally, the study of ``part-aware attention networks"~\cite{yan2019learning} found that occluding one facial component during training increased KV accuracy. On the other hand, we consider training based on facial features to reduce computational cost. Considering all these notions, we propose our method based on graph neural concepts, generally depicted in Fig.~\ref{fig:propose-method}, and explain its steps as follows:

\begin{figure*}[pos=htbp]
	\begin{center}
		\includegraphics[width=\textwidth]{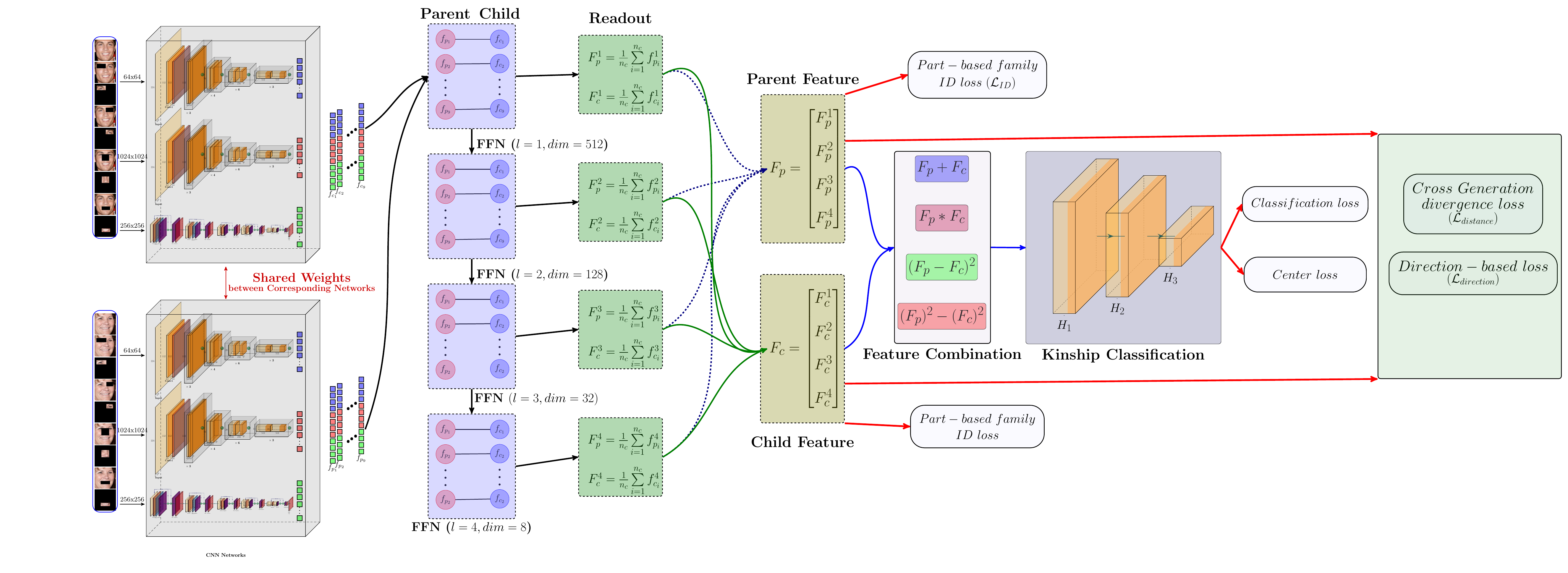}
		\caption{The General Diagram of the Proposed Method. The primary CNNs, ResNet-18 and ERN, for each kinship dataset are trained before being fed to the GNN layers, except for VGGFace2. Information exchange happens in FNN layers. In the readout phase, from parent and child nodes, a corresponding feature vector is constructed. The various combinations of either parental or child feature vectors are constructed and fed into the classification module.}
		\label{fig:propose-method}
	\end{center}
\end{figure*}

\begin{figure*}[pos=htbp]
	\begin{center}
		\includegraphics{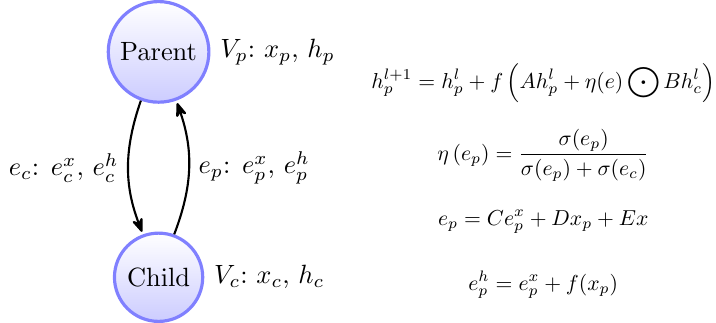}
		\caption{A concise representation of Residual Gated Graph ConvNets for two nodes}
		\label{fig:vertex}
	\end{center}
\end{figure*}

\begin{figure*}[pos=htbp]
	\begin{center}
		\includegraphics[width=\textwidth]{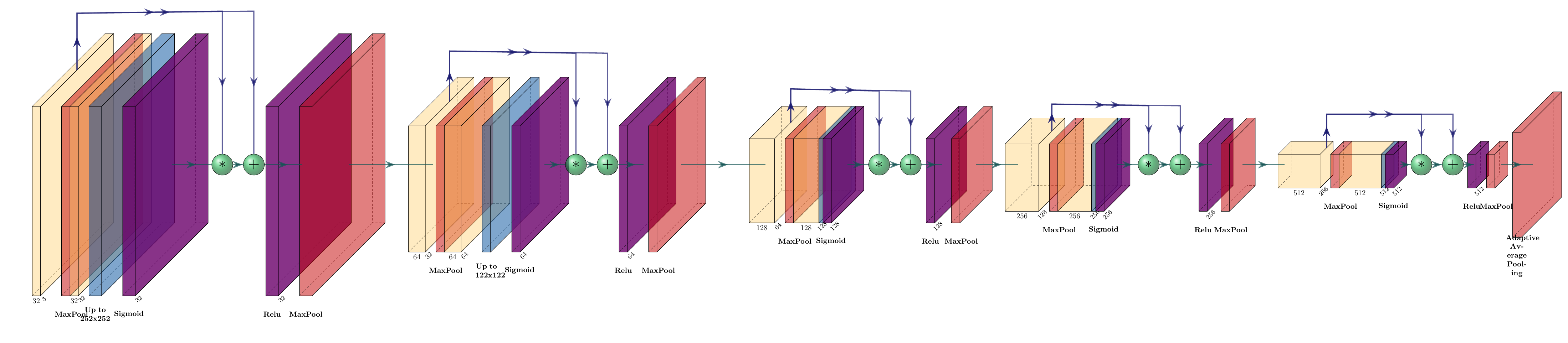}
		\caption{The block diagram of the Extended Residual Network}\label{fig:residual-network}
	\end{center}
\end{figure*}

\begin{figure*}[pos=htbp]
	\begin{center}
		\includegraphics[width=\textwidth]{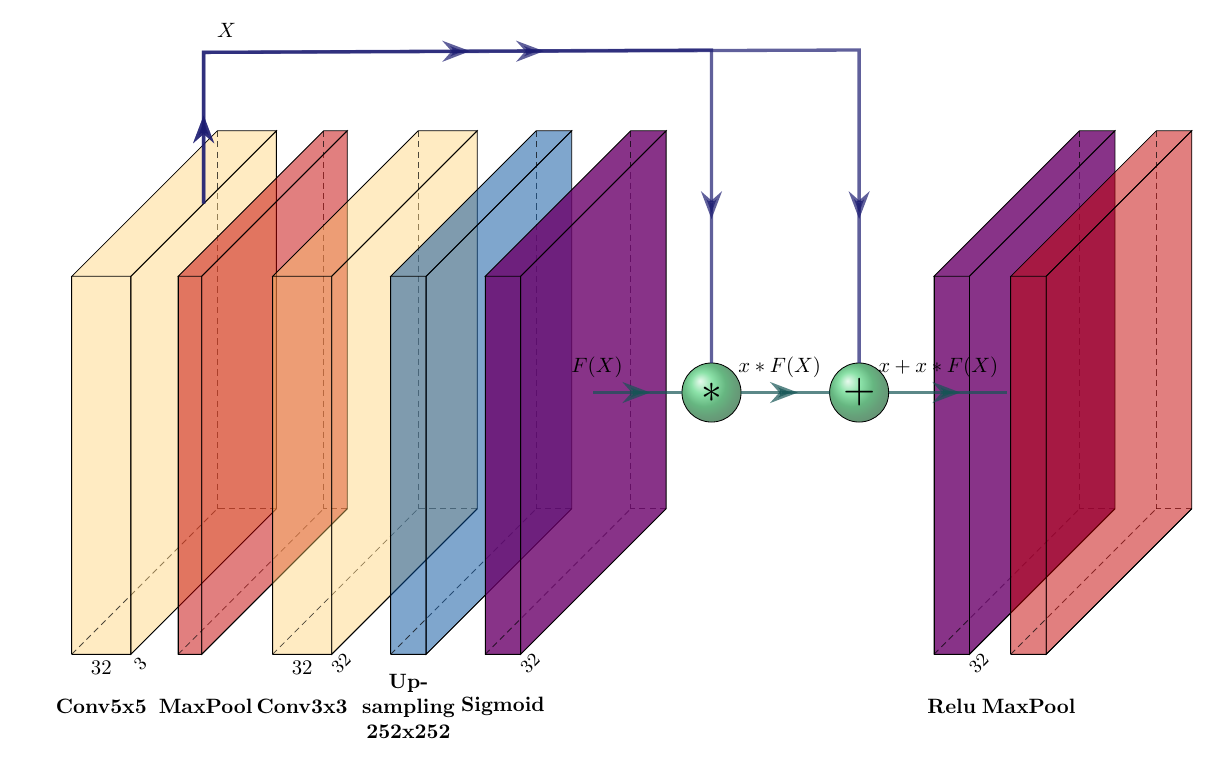}
		\caption{One block of the Extended Residual Network (ERN)}
		\label{fig:one-block-ern}
	\end{center}
\end{figure*}

\textbf{First}, CNNs are trained on face images to learn facial features.
\textbf{Second}, facial landmarks are detected, and then the face components are extracted. We obtain the embeddings of those patches using CNNs trained beforehand. 
\textbf{Third}, various graphs are composed of corresponding patches between the parent and child face components. For example, a graph of two nodes and a bidirectional edge between a parent's right eye and a child's right eye is constructed. Since we construct this graph from two corresponding components in the parent and child images, we call it,~\textbf{a paired graph}. 
Paired graphs are 9 graphs, of which 5 consist of the whole face images, right eyes, left eyes, noses, and mouths, obtained from embeddings of the parent and child's face images. Four other graphs are built from patches that do not contain a face component in the facial image. 
To explain more, if the right eye of the face image is blocked, we will have a face image without the right eye. This approach can construct four images. Then, for each of these four images of a parent and a child, four graphs are built. Eventually, these nine paired graphs constitute a forest. 

\textbf{Finally}, we give this forest to a graph neural network, the Forest Neural Network, which performs GNN-like operations to exchange information between nodes and model family relationships.

\begin{figure}[pos=htbp]
	\begin{center}
		\includegraphics{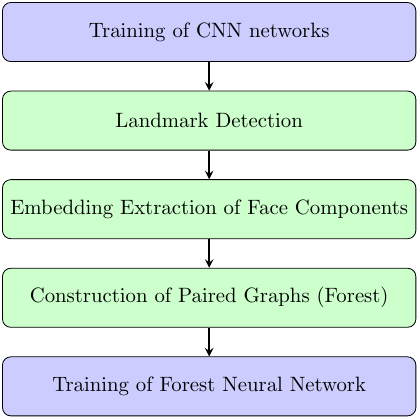}
		\caption{The steps of the proposed algorithm}
		\label{fig:flowchart}
	\end{center}
\end{figure}

The steps of the proposed algorithm are enumerated to clarify its understanding and depicted in Fig.~\ref{fig:flowchart}. 

\begin{enumerate}
	\item Training CNNs to obtain facial features
	\item Landmark Detection
	\item Embedding Extraction of Patches
	\item Construction of Paired Graphs, i.e, a Forest
	\item Training Paired Graphs by Forest Neural Network
\end{enumerate}

\subsection{Training CNN networks}
Despite transformers having shown promising results, especially on large datasets, CNNs still have comparative results to transformers, can generalize more, and are well-suited to limited data due to the small size of their network~\cite{mauricio2023comparing, talib2024performance}. 

Additionally, various CNNs can be seen as expert systems and employed to enhance our knowledge about data and obtain different embedding vectors. Although various networks of the same task generally attempt to learn the equivalent data distributions, they have slightly different representations since they have different performances. This slight difference is fruitful to increase the performance of small datasets. Hence, we utilize various CNNs, such as ReseNet-18, which are used in different computer vision tasks, and the Extended Residual Network, described in~\ref{subsection:ERN}.

A significant concern when applying CNNs is that there is only one sample per identity in KinFaceW-I and II. Since these datasets have few samples, data augmentation is required to train neural networks. Therefore, we apply various augmentation techniques and masked images, such as complete face images without the right eye, left eye, nose, and mouth, to train networks.

In addition, high-resolution facial images can improve accuracy by providing much information about faces. Since we do not have high-resolution facial images, one solution is to apply GANs to produce them.
Moreover, employing GANs has another advantage because they produce images in the same domain and remove differences between parent and child images taken under different conditions.

When choosing GANs, perceptual metrics such as FID~\cite{heusel2017gans} and NIQE~\cite{mittal2012making} can help identify a good candidate. Among the cutting-edge GANs for super-resolution, we evaluate them based on FID and NIQE scores, visually inspect their outputs, and select GFP-GAN~\cite{wang2021towards}. This blind face-restoration method produces 256×256 images. After that, the ResNet is retrained for this new dataset to gain new embeddings.

\subsection{Extended Residual Network (ERN)}
\label{subsection:ERN}
We redesign the neural network, applied in~\cite{yan2019learning,zhu2023distance}, to produce higher-dimensional feature vectors. We consider the idea that neural networks can produce rich feature representations by shrinking the width and height of their inputs and increasing depth via a set of kernels. When we need a feature vector of dimension 128, one approach is to set the layer's feature map depth to 128 and apply average pooling to all other dimensions. We can set various intermediate depths from 3 input channels. However, due to computational efficiency and predictable memory or parameter growth, powers of two (32, 64, 128) have been chosen, as discussed in the channel-doubling fabrics section of ``Convolutional Neural Fabrics" ~\cite{saxena2016convolutional}.

The attention mechanism in~\cite{yan2019learning,zhu2023distance} is used to highlight the local parts that play a significant role in kinship verification. An ERN block consists of a convolution (kernel=5×5), max-pooling, another convolution (kernel=3×3), up-sampling, and sigmoid layers applied to the combination of paired parent-child images as shown in Fig.~\ref{fig:one-block-ern}.

These consecutive operations produce a feature map with values in the range [0, 1]. This feature map is multiplied by the original to highlight the significant parts. Finally, the original feature map is added to counteract the disappearance of some features, as in all residual networks. 
The paper~\cite{zhu2023distance} utilized a residual block attention module for extracting shared features. However, the parent's and the child's face images are fed to the module separately. Then, different combinations of feature maps were used for the classification. 

Despite the module's use in~\cite{yan2019learning,zhu2023distance}, we employ this attention mechanism to obtain richer representations of the entire images in the dataset.
We must make adjustments to produce an embedding vector consistent with other CNNs if we use three layers of the attention block, as in~\cite{zhu2023distance}. This results in a high-dimensional feature map, making the entire model enormous. We indicate the specifications of the convolutional and up-sampling layers in three blocks in Table~\ref{table:conv-128}.

\begin{table*}[htbp]
	\caption{Convolutional layers and up-sampling size of all blocks to produce 128-dimensional vectors. The padding and stride for all convolutional layers are 0 and 1, respectively.}
	\label{table:conv-128}
	\begin{tabular}{l|ccc}
		\begin{tabular}[c]{@{}c@{}}Convolution\\ Layer\end{tabular}
		& \begin{tabular}[c]{@{}c@{}}In\\ channel\end{tabular} 
		& \begin{tabular}[c]{@{}c@{}}Out\\ channel\end{tabular} 
		& \begin{tabular}[c]{@{}c@{}}Kernel\\ Size\end{tabular} 
		\\ \toprule
		The 1$^{st}$ conv of block 1 & 3          & 32          & (5,5)       \\ 
		The 2$^{nd}$ conv of block 1 & 32         & 32          & (3,3)       \\ 
		The 1$^{st}$ conv of block 2 & 32         & 64          & (5,5)       \\ 
		The 2$^{nd}$ conv of block 2 & 64         & 64          & (3,3)       \\ 
		The 1$^{st}$ conv of block 3 & 64         & 128         & (5,5)       \\ 
		The 2$^{nd}$ conv of block 3 & 128        & 128         & (3,3)       \\ 
	\end{tabular}
	\begin{tabular}{c|c}
		Up-sampling & Size      \\ \hline
		Block 1           & (60,60) \\ \hline
		Block 2           & (26,26) \\ \hline
		Block 3           & (9,9)   \\ \hline
	\end{tabular}
\end{table*}

One way to reduce the feature map's dimension is to use a pooling layer at the end of all the attention blocks, such as an adaptive average pooling layer. Thus, we can use an adaptive pooling layer at the end of each block to produce a feature vector that differs from the others. Therefore, these blocks are changed to gain a feature vector of size 512. 

Due to the attention block's usage, we resize the input image to 256×256, as consecutive convolution operations reduce the last two dimensions of the feature map to minimal values. The specifications of the convolutional and up-sampling layers in all blocks are shown in Table~\ref{table:conv}. For brevity, we do not specify the padding and stride of all convolution layers individually; they are set to 0 and 1, respectively, for all blocks.

\begin{table*}[htbp]
	\caption{Convolutional layers and up-sampling size of all blocks to produce 512-dimensional vectors. The padding and stride for all convolutional layers are 0 and 1, respectively.}
	\label{table:conv}
	\begin{tabular}{l|ccc}
		\begin{tabular}[c]{@{}c@{}}Convolution\\ Layer\end{tabular}
		& \begin{tabular}[c]{@{}c@{}}In\\ channel\end{tabular} 
		& \begin{tabular}[c]{@{}c@{}}Out\\ channel\end{tabular} 
		& \begin{tabular}[c]{@{}c@{}}Kernel\\ Size\end{tabular} 
		\\ \toprule
		The 1$^{st}$ conv of block 1 & 3          & 32          & (5,5)       \\ 
		The 2$^{nd}$ conv of block 1 & 32         & 32          & (3,3)       \\ 
		The 1$^{st}$ conv of block 2 & 32         & 64          & (5,5)       \\ 
		The 2$^{nd}$ conv of block 2 & 64         & 64          & (3,3)       \\ 
		The 1$^{st}$ conv of block 3 & 64         & 128         & (5,5)       \\ 
		The 2$^{nd}$ conv of block 3 & 128        & 128         & (3,3)       \\ 
		The 1$^{st}$ conv of block 4 & 128        & 256         & (5,5)       \\ 
		The 2$^{nd}$ conv of block 4 & 256        & 256         & (3,3)       \\ 
		The 1$^{st}$ conv of block 5 & 256        & 512         & (5,5)       \\ 
		The 2$^{nd}$ conv of block 5 & 512        & 512         & (3,3)      
	\end{tabular}
	\begin{tabular}{c|c}
		Up-sampling & Size      \\ \toprule
		Block 1           & (252,252) \\ 
		Block 2           & (122,122) \\ 
		Block 3           & (57,57)   \\
		Block 4           & (24,24)   \\ 
		Block 5           & (8,8)     \\ 
	\end{tabular}
\end{table*}

The schematic of the Extended Residual Network~(ERN) is demonstrated in~\ref{fig:residual-network}.

\subsection{Landmark Detection}\label{subsec:propose-landmark-detection-section} 
To locate the face components, it is necessary to find the landmark positions in the face image. Some famous landmark detections include MTCNN~\cite{zhang2016joint} and DLIB~\cite{kazemi2014one}. However, one of the state-of-the-art landmark detections is SPIGA~\cite{Prados-Torreblanca_2022_BMVC}, which obtained the highest results in different benchmarks. The success of SPIGA lies in its ability to learn both the global representation of the facial structure (geometrical information) through a cascade of graph attention networks and the local representation using the CNN architecture. The global representation was indicated by a fixed and fully connected weight graph, and its nodes were landmarks. SPIGA was evaluated on various hard subsets of the WFLW data set~\cite{wu2018look}, such as pose, illumination, make-up, occlusion and blur. Based on our observation on KinFaceW-I and KinFaceW-II~\cite{lu2013neighborhood}, all landmarks of the images can be fully identified using SPIGA. 


\subsection{Embedding Extraction of Patches}~\label{subsec:patch-extraction}
Four bounding boxes of the right and left eye, nose, and mouth are extracted by the coordinates found in Section Landmark Detection~\ref{subsec:propose-landmark-detection-section}.
By landmark points, we can specify a bounding box around the facial components.
The minimum and maximum points on the x- and y-axes of the landmarks define the bounding box. For example, the right eyebrow and right eye landmarks represent the right eye box.
In addition, we extract the full face image without the right eye bounding box and refer to this image hereafter as the face image without the right eye. In this approach, we generate face images without a left eye, nose, or mouth, and in total, obtain 9 images per facial image.
Then, the trained CNNs are applied to obtain embedding vectors for all 9 images.

\subsection{Forest Graph Construction and Training}
Based on the hypothesis mentioned at the beginning of  Section~\ref{sec:proposed-method}, paired graphs between parent and child images are being constructed. These paired graphs construct a forest of graphs.

For training the forest graph, the GCN algorithm is applied to each paired graph separately. Then, at the end of the epoch, a mean aggregation is calculated among all parent nodes from all nine graphs. The obtained feature is considered the parent's feature vector. By this approach, we can obtain a feature vector for the child.

\subsection{Deatails of Training Forest Graph Network}~\label{sec:fnn}
Given the rich information of facial components and the hypothesis that people assess others' similarity through them, we recommend the Forest Neural Network (FNN) to process this information for family relationships.
It has operators and computations similar to those in graph neural networks, but it uses separate graphs. This paper uses a kind of GCN known as Residual Gated Graph Convnets~\cite{bresson2017residual} shown in Fig.~\ref{fig:vertex}. GCN is the basic type of GNN and performs message passing and neighborhood aggregation using the same operations as GNN, except that it uses convolution to propagate information across nodes.
Since every graph in our FNN contains only two parent and child nodes, the message-passing mechanism between parent and child nodes of every graph in FNN is simplified as follows:

The nodes and edges of the graphs have their own feature vectors and are updated iteratively according to their respective equations. Every node’s initial feature vector  is denoted as X, and after updating, it is referred to as the hidden feature vector, \textbf{h}, or the hidden representation.
The hidden representations of parent and child nodes are denoted as \textbf{$h_p$} and \textbf{$h_c$}, respectively. 

The initial feature vector of each edge is denoted \textbf{$e^x$} and after updating, it is referred to as \textbf{$e^h$}. 

The hidden representations, \textbf{$h_p$} and \textbf{$h_c$}, for parent and child nodes are updated by Eq.~\ref{eq:hidden-vertex}. All equations are written for the parent nodes at layer $l+1$, and the child nodes are alike.
\begin{equation}
	h^{\l+1}_{p} = h^{\l}_{p}+ f\left ( Ah^{\l}_{p} + \eta (e)\bigodot Bh^{\l}_c \right)
	\label{eq:hidden-vertex}
\end{equation}
In Eq.~\ref{eq:hidden-vertex},~$h^0_p=x_p$, $h^0_c=x_c$; $h^l_p$ and $h^l_c$ are the current parent and child representations; the matrices A, and B are learnable parameters, and f is an activation function like ReLU. The operation $\bigodot$ is the element-wise (Hadamard) product.
The gate term~$\eta (e_p)$ weighs the amount of information exchanged between the parent node and its neighbor, the child node. Namely, it controls the amount of information passed from the neighboring node and calculated in Eq.~\ref{eq:gate-term}.
The residual connection $h^{\l}_{p}+...$ helps preserve information from previous layers.
\begin{equation}
	\eta \left ( e_p\right)=\frac{\sigma (e_p)}{\sigma (e_p)+\sigma(e_c)}~\label{eq:gate-term}
\end{equation}
The output of the sigmoid function~$\sigma (.)$ is between 0 and 1. If $\sigma (e_p)$ is large, then the parent’s information dominates.
The edge representation $e_p$, is defined and updated by Eq.~\ref{eq:edge} :
\begin{equation}
	e_p=Ce^x_p+Dx_p+Ex
	~\label{eq:edge}
\end{equation}
The matrices C, D, and E are learnable parameters.
This Eq.~\ref{eq:edge} uses the parent node information, i.e., $x_p$, neighboring information, x, and its previous edge information, $e^x$.
\begin{equation}
	e^h_p=e^x_p+f(x_p)
	~\label{eq:edge2}
\end{equation}
Eq.~\ref {eq:edge2} illustrates the update of the edge presentation and is similar to the residual update of the node representation. The next edge representation, $e^h_p$, is obtained by the previous edge representation, $e^x$, and the output of the nonlinear function on that node representation, $f(x_p)$. All updates of nodes and edges on the nine graphs are performed based on Eq.~\ref{eq:hidden-vertex} to \ref{eq:edge2}.

\textit{Readout phase}: the final state of FNN is the aggregation function, i.e., mean, operated to all parent nodes of the nine graphs. In other words, the mean of all parent nodes in the forest represents a feature vector for the parent in that layer, $F_p^l$ as written in Eq.~\ref{eq:single-layer-mean}. The $n_{nodes}$ is the number of parent or child nodes.
\begin{equation}
	F^l_p= \frac{1}{n_{nodes}} \sum_{1}^{n_{nodes}}f^l_{p_{i}}
	~\label{eq:single-layer-mean}
\end{equation}
Then, all these mean vectors of parent nodes from all layers are concatenated together as the final feature vector for the parent, $F^p$, depicted in Eq.~\ref{eq:parent-feature-vector}.
\begin{equation}
	F^p = \begin{bmatrix}
		F_p^1 \\[1em]
		F_p^2 \\[1em]
		F_p^3 \\[1em]
		F_p^4 
	\end{bmatrix}
	~\label{eq:parent-feature-vector}
\end{equation}
The child's final feature vector, $F^c$, is also obtained alike.
The final parent and child feature vectors are sent to the next block for the classification phase.

\subsection{Kinship Classifier}
\textit{Feature Combination}: different combinations of two feature vectors can be helpful because each form of combination resembles a different metric. In addition, if we scrutinize various distance functions, we can observe that there is a form of addition, difference, and power at the core of all of them.
Hence, before feeding features of the $i^{th}$ parent-child pair extracted from~\ref{sec:fnn} to a classifier, different combinations of features formulated in Eq.~\ref{eq:fuse} are computed and concatenated together like~\cite{zhu2023distance}.  
\begin{equation}
	F^{p,c}_i= \begin{pmatrix}
		\left( F^p_i -F^c_i \right)^2  \\[1em]
		\left(F^p_i +F^c_i\right)      \\[1em]
		\left(F^p_i \ast F^c_i \right)  \\[1em]
		\left( (F^p_i)^2 -(F^c_i)^2 \right)	
	\end{pmatrix}
	~\label{eq:fuse}
\end{equation}
\textit{Classifier}: We need three linear layers because the first layer accepts the diverse combinations of features and maps them to a lower dimension. Another layer, i.e., the last layer of the classifier, is needed to map the dimensions to one or two neurons and infer whether these features indicate a family relationship. The middle layer is required due to the center loss. Therefore, we decrease the input dimensions in three consecutive layers: first, the dimension of feature combinations; second, the dimension required by the center loss; and finally, the dimension of classification.

\subsection{Loss Functions}
We employ various loss functions to reduce network error from different perspectives, as in the work~\cite{zhu2023distance}. Additionally, we have added the center loss to other losses due to its effectiveness in improving network accuracy. Generally, we use four groups of loss functions:

\begin{enumerate}
	\item Pair- and identity-based losses
	\begin{itemize}
		\item Kinship pair classification loss ($\mathcal{L}_{kin}$)
		\item part-based family-ID loss ($\mathcal{L}_{ID}$)
	\end{itemize}    
	\item Distance-based loss
	\begin{itemize}
		\item cross-generation divergence loss ($\mathcal{L}_{distance}$)
	\end{itemize}
	\item Direction-based loss
	\begin{itemize}
		\item direction consistency loss ($\mathcal{L}_{direction}$)
	\end{itemize}
	\item Center-based loss
	\begin{itemize}
		\item center loss (encouraging intra-class compactness, $\mathcal{L}_{center}$)
	\end{itemize}
\end{enumerate}

Their combination of losses is formulated in Eq.~\ref{eq:loss-combination}.
\begin{equation}
	\mathcal{L}=\omega_{kin}\mathcal{L}_{kin} + \omega_{ID}\mathcal{L}_{ID} + \omega_{distance}\mathcal{L}_{distance} + \omega_{direction}\mathcal{L}_{direction} + \omega_{center}\mathcal{L}_{center}
	~\label{eq:loss-combination}
\end{equation}

\subsubsection{Kinship Pair Classification Loss~($\mathcal{L}_{kin}$)}
Because the classifier has only one neuron in the last layer, we use the binary cross-entropy loss function to measure the network's classification error. For the sake of numerical stability, it is better to use BCEWithLogitsLoss, combining a sigmoid layer and the Binary Cross Entropy as formulated in Eq.~\ref{eq:bcewithlogit}. 

\begin{equation}
	\mathcal{L}_{kin} = -\frac{1}{N} \sum_{i=1}^{N} \left[ y_i \log\sigma(\hat{y}_i) + (1 - y_i) \log(1 - \sigma(\hat{y}_i)) \right]
	~\label{eq:bcewithlogit}
\end{equation}
Additionally, $y_i$ denotes the class label and $y_i=1$ if the parent and child in the $^i$th image pair have a kinship relation, otherwise $y_i=0$. We also denote the predicted label as $\hat{y}_i$.

\subsubsection{Part-based Family-ID Loss~($\mathcal{L}_{ID}$)}
Family members have the same identity as the family. Therefore, we can assign a specific label to all family members. Hence, we assign the same label to images that share the kinship relation; for example, the label $z_i$ to the two images in the $i^{th}$ image pair, the child and the parent.
On the other hand, it is shown in~\cite{zhu2023distance, sun2019perceive} that splitting a feature vector into parts with the same identity improves the network's robustness. However, the number of optimized divisions must be explored. This loss is defined in Eq.~\ref{eq:family-id-loss}.
\begin{equation}
	\mathcal{L}_{ID} (F_i, z_i) = -\frac{1}{N}\frac{1}{M} \sum_{i=1}^{N} \sum_{j=1}^{M} z_i \log p(z_i|F_{i,j})
	~\label{eq:family-id-loss}
\end{equation}

Where the number of facial images and their parts are N and M, respectively, the $F_i$ and $F_{i,j}$ are the feature vector, and its j-part of the feature vector of the $i^{th}$ image (can be either the parent or child), respectively.

\subsubsection{Distance-based or Cross generation divergence decreasing Loss~($\mathcal{L}_{distance}$)}
It is beneficial to reduce the distance between parent and child feature vectors, since we can classify closely related feature vectors more easily in the next stage. This loss $\mathcal{L}_{cross}$ in Eq.~\ref{eq:cross-loss} reduces the gap between the parent and child features in the graph part of the network before the classification module.
\begin{equation}
	\mathcal{L}_{cross} = \omega_1 \mathcal{L}_{pos} + \omega_2 \mathcal{L}_{neg}
	~\label{eq:cross-loss}
\end{equation}

\begin{equation}
	\mathcal{L}_{pos} = \frac{1}{n} \sum_{i=1}^{n} y_i * \left\| F_i^p - F_i^c \right\|_2^2
	~\label{eq:cross-pos-loss}
\end{equation}

\begin{equation}
	\mathcal{L}_{neg} = \frac{1}{n} \sum_{i=1}^{n} (1-y_i) * \left\| F_i^p - F_i^c \right\|_2^2
	~\label{eq:cross-neg-loss}
\end{equation}
Where weights, $\omega_1$ and $\omega_2$, denote the importance of reducing the gap among positive or negative samples. The $F_i^p$ and $F_i^c$  are the parent and child feature vectors of the $i^{th}$ paired images.

\subsubsection{Direction-based Loss~($\mathcal{L}_{direction}$)}
It is worthwhile to use the angular distance between samples to determine whether they are in the same direction. One of the ways to measure the angle among features is the cosine distance defined in Eq.~\ref{eq:cosine-distance}, which takes two features of the parent ($F_i^p$) and child ($F_i^c$) in the $i^{th}$ image pair: 
\begin{equation}
	\mathrm{Cos}(F_i^p, F_i^c) = \frac{F_i^p . F_i^c}{\left\| F_i^p \right\|_2 \left\| F_i^c \right\|_2}
	~\label{eq:cosine-distance}
\end{equation}
It will be one if they have a zero angle in the same direction. If they are in opposite directions, the angle is $180^{\circ}$, then the distance is -1. If their angle is between zero and $180^{\circ }$, their distance is between -1 and +1.

This direction-based loss Eq.~\ref{eq:loss-direction} measures the angular distance between positive and negative samples so that positive samples possess the highest similarity and negative ones possess the least similarity. Hence, similar pairs must have the label $y_i=+1$, and dissimilar ones must have the label $y_i=-1$.

\begin{equation}
	\mathcal{L}_{direction}\left( cos(F_i^p, F_i^c), y_i \right) = \frac{1}{n} \sum_{i=1}^{n} \left( cos(F_i^p, F_i^c) - y_i \right)^2
	~\label{eq:loss-direction}
\end{equation}

In Eq.~\ref{eq:loss-direction}, we do not use any margin or clip the values.

\subsubsection{Center-based Loss~(Center Loss, $\mathcal{L}_{center}$)}
This loss aims to contract the samples within the same class in the feature space and enhance the network's discriminatory power. It learns the centers in the feature map space for each class and penalizes samples that are further from these centers~\cite{wen2016discriminative}. Except for the binary cross-entropy loss, other losses aim to reduce the error in the network's innermost layers, especially those of the forest neural network. By using this center loss, we can control the discriminatory power of the kinship classification module and place greater pressure on its layers to produce more compact features. Its formulation is represented in Eq.~\ref{eq:center-loss}.

\begin{equation}
	\mathcal{L}_{center} = \frac{1}{2} \sum_{i=1}^{N} \left\| H_i - C_{y_i} \right\|_2^2
	~\label{eq:center-loss}
\end{equation}
Where $H_i \in \mathbb{R}^d$ denotes the $i^{th}$ output of the hidden layer in the kinship classification module, centers $C_{y_i}\in \mathbb{R}^d$ possess the same dimension as $H_i$ and are randomly initialized. Their number equals the class number; here are two classes. These centers are learned during the training phase.

\subsection{Fusion of Losses}
Each loss plays a role in guiding the network through its various perspectives. For example, the cross-entropy loss aims to minimize misclassification error. Moreover, the center loss attempts to guide the feature map to produce features near class centers. The direction-based loss considers the angle between features. In addition, losses such as the cross-generation loss minimize the family identity of images. The scale of these losses is significant because large values can dominate smaller ones. Therefore, various combinations of these losses can impact the network's performance. 

Additionally, a loss may not contribute to performance until after some epochs of training, as in the case of the center loss. It is gradually incorporated over the training epochs because it can have a positive impact when the feature maps of two classes are somewhat separated. Thus, we introduce the temperature rate to handle this involvement as formulated in Eq.~\ref{eq:fusion-loss}.

\begin{equation}
	\mathcal{L}_{fusion} = \omega_0({\alpha^t})\cdot\mathcal{L}_{center} + \sum_{i=1}^{M} \omega_i\mathcal{L}_{i} 
	~\label{eq:fusion-loss}
\end{equation}
The temperature rate ${\alpha^t}$ at the early stage of training is small and increases in every epoch $t$. Every $\omega_i$ is a constant number determined by its corresponding loss. The temperature base $\alpha$ is also a constant number. 
The pseudo-code for the proposed algorithm, which investigates how we apply FNN and losses, is shown in Alg.~\ref{alg:train}.


\begin{algorithm}
	\caption{\textbf{The Proposed FNN For Kinship Verification}:}
	\label{alg:train}

	\begin{algorithmic}[1]
		\Require
		\Statex $\triangleright$~ $x_1$ and $x_2$ are parent and child images
		\Statex	$\triangleright$ $F_p$, $F_c$ are Parent and child Features from FNN.
		\Statex $\triangleright$ H is the feature of MLP for the center loss.
		\Statex $\triangleright$ n\_part determines the number of division for features
		
		\Ensure
		\Statex $\triangleright$ epoch\_loss shows the loss in every epoch
		\Statex $\triangleright$ epoch\_trn\_acc denotes the current accuracy in epoch
		\Statex
		
		\State $\text{n\_data} \gets 0$
		\State $\text{epoch\_loss} \gets 0$
		\State $\text{epoch\_train\_acc} \gets 0$

		
		
		\ForAll{$\text{(x1, x2, labels, parent\_ids, child\_ids)}$ \\
			\hspace{2em} $ \in \text{dataloader}$}
		
		\State $\text{Move all data to GPU}$
		\State $\text{ground\_truth} \gets \text{flatten(labels)}$
		\State $\text{target} \gets \text{concat}(parent\_ids, child\_ids)$
		
		
		\State $~$
		\State $\text{scores, Fp, Fc, family\_id, H} \gets \text{\textbf{FNN\_Model}(x1, x2)}$
		
		\State $\text{predictions} \gets (\text{batch\_scores} > 0.5)$
		
		\State $\text{epoch\_trn\_acc} \gets \text{epoch\_trn\_acc}+...$
		\State $\text{sum of all samples correctly classified}$
		\State $~$
		\State $\text{n\_data} \gets \text{n\_data} + \text{n\_samples\_in\_batch}$
		
		\State $\text{f} \gets \text{concat}(F_p, F_c)$
		
		
		\State $\text{p1, p2} \gets \text{Compute positive sample features}$
		
		\State $\text{n1, n2} \gets \text{Compute negative sample features}$
		
		\State $\text{c\_similarity} \gets \text{cosine\_similarity}(Fp, Fc)$
		
		\State $\text{Compute Losses:}$
		\State $\mathcal{L}_1 \gets \text{BCE(batch\_scores, ground\_truth)}$	
		\State $\mathcal{L}_2 \gets \text{MSE(p1, p2)}$
		\State $\mathcal{L}_3 \gets \text{MSE(n1, n2)}$
		\State $\mathcal{L}_4 \gets \text{MSE(c\_similarity, ground\_truth)}$
		
		\For{$i \gets 1 \text{ to n\_part}$}
		\State $\mathcal{L}_5 \gets \mathcal{L}_5 + \text{CrossEntropy(family\_id, target)}$
		\EndFor
		
		\State $\mathcal{L}_6 \gets \text{CenterLoss~(H, ground\_truth)}$
		
		\State $\mathcal{L}_{fusion} \gets \omega_0\cdot({\alpha^t}) \cdot \mathcal{L}_{center}$
		\For{$i \gets 1$ to $6$}
		\State $\mathcal{L}_{fusion} \gets \mathcal{L}_{fusion} + \omega_i \cdot \mathcal{L}_i$
		\EndFor
		
		\State $\text{set gradients of parameters in optimizers to zero.}$
		\State $\mathcal{L}_{fusion}.backward()$
		\State $\text{optimizer.step()}$
		
		\ForAll{$\text{param} \in \text{Center\_loss.parameters()}$}
		\State $\text{param.grad.data} \gets \text{param.grad.data} \cdot (\omega_0{\alpha^t})^{-1}$
		\EndFor
		\State $\text{Center\_optimizer.step()}$
		\State $\text{epoch\_loss} \gets \text{epoch\_loss} + \mathcal{L}_{fusion}$
		\EndFor
		
		\State $\text{epoch\_loss} \gets \text{epoch\_loss} / (\text{iter} + 1)$
		\State $\text{epoch\_trn\_acc} \gets \text{epoch\_train\_acc} / \text{n\_data}$
		\State \Return $\text{epoch\_loss, epoch\_trn\_acc}$
	\end{algorithmic}
\end{algorithm}

\section{Experiment}
\label{sec:experiment}
This section mentions datasets and experimental configurations. First, we introduce public datasets for kinship verification, then explain the implementation details and experiments for CNNs and our FNN.
\subsection{Dataset}
\label{subsection:dataset}
We investigate the effectiveness of our proposed method on KinFaceW-I and KinFaceW-II~\cite{lu2013neighborhood}, two benchmark datasets on which most kinship verification algorithms have conducted their experiments, as they include a standard protocol that ensures fair comparisons. These datasets consist of a single sample per identity, which makes the problem harder. 


These datasets provide unconstrained facial images of parents and their children. They comprise four types of kinship relationships: Father-Son (FS) and Father-Daughter (FD), Mother-Son (MS), and Mother-Daughter (MD). KinFaceW-I owns 156, 134, 116, and 127 image pairs for each kinship type. KinFaceW-II comprises 250 image pairs per kinship relationship. These datasets have a standard configuration of five folds of paired parent-child images. Each fold consists of a balanced set of paired images, 50 positive and 50 negative samples. This configuration obligates researchers to conduct experiments in the same way and to ensure a fair comparison. 


\subsection{Configuration of CNN Networks}
\begin{figure}[pos=htbp]
	\begin{center}
		\includegraphics[trim={4.9cm 12.4cm 3.2cm 12.4cm},clip, width=\textwidth]{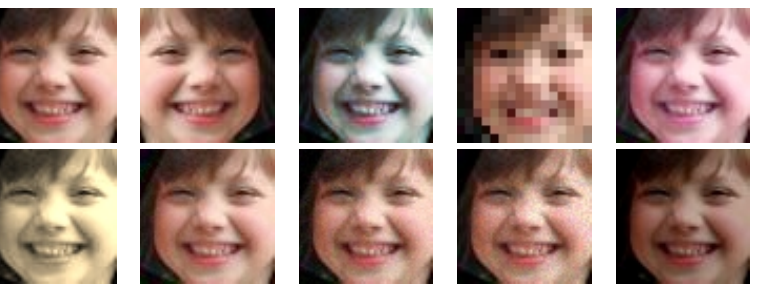}
		\caption{The original child image and the other augmented pictures are arranged in the following sequence, from left to right and top to bottom: Original, Flip, Equalize, Downscale, Hue Saturation, Random Rain, Sepia, ISO Noise, Gaussian Noise, Multiplicative Noise, Random Brightness Contrast, and Shift Scale Rotate.}
		\label{fig:daughter_augmentation}
	\end{center}
\end{figure}

We describe the implementation details for extracting embedding features from ResNet-18 and ERN here. Since only one sample per identity is available in KinFaceW-I and KinFaceW-II, and we need at least one sample for training and another for testing, we need to increase the sample size through augmentation. 
We employ the various augmentation techniques, listed in~Table~\ref{table:augmentation}, from the Albumentation package~\cite{info11020125} and display the photos produced in Fig~\ref{fig:daughter_augmentation}. 

\begin{table*}[]
	\caption{List of various augmentation techniques and their parameters used to train CNNs.}
	\resizebox{\textwidth}{!}{
		\begin{tabular}{ll|ll}
			
			\textbf{Augmentation Technique}    & \textbf{Parameters}                                                                                                                                       & \textbf{Augmentation Technique} & \textbf{Parameters} \\ \toprule
			
			Flip~(Horizontal)             & default & Sepia                      & default       \\ \hline                                                                                                
			
			Random Rotate 90 & default  & ISO Noise                  & \begin{tabular}[c]{@{}l@{}}  intensity=(0.1, 0.5), color\_shift=(0.01,0.05)\end{tabular} \\ \hline
			
			Equalize         & mode='cv', by\_channels=True & Gaussian Noise             & var\_limit=(10.0, 50.0) \\ \hline
			
			Downscale       & \begin{tabular}[c]{@{}l@{}} scale\_min=0.25, \\ scale\_max=0.25, interpolation=0\end{tabular}                                                                                   & Multiplicative Noise       & \begin{tabular}[c]{@{}l@{}} multiplier=(0.9, 1.1), \\ per\_channel=True, elementwise=True\end{tabular}                                                                                                                                         \\ \hline
			Hue Saturation  & \begin{tabular}[c]{@{}l@{}}hue\_shift\_limit=(-20, 20),  \\ sat\_shift\_limit=(-30, 30), \\ val\_shift\_limit =(-20, 20)\end{tabular}                                           & Random Brightness Contrast & \begin{tabular}[c]{@{}l@{}} brightness\_limit=(-0.2, 0.2), \\ contrast\_limit=(-0.2, 0.2), \\ brightness\_by\_max=True\end{tabular}                                                                                                         \\ \hline
			Random Rain      & \begin{tabular}[c]{@{}l@{}}slant\_lower=-10, slant\_upper=5, \\ drop\_length=10, drop\_width=1, \\ drop\_color=(0, 0, 0), blur\_value=2, \\ brightness\_coefficient=0.7,\end{tabular} & Shift Scale Rotate         & \begin{tabular}[c]{@{}l@{}}rotate\_limit=(-45, 45),  \\ shift\_limit=(-0.0625, 0.0625), \\ scale\_limit=(-0.1, 0.1), \\ interpolation=cv2.INTER\_NEAREST, \\ border\_mode=cv2.BORDER\_CONSTANT, \\ rotate\_method= 'largest\_box'\end{tabular} \\ 
		\end{tabular}
	}
	\label{table:augmentation}
\end{table*}

\begin{figure}[pos=htbp]
	\begin{center}
		\includegraphics[trim={0cm 25cm 10.5cm 0cm},clip]{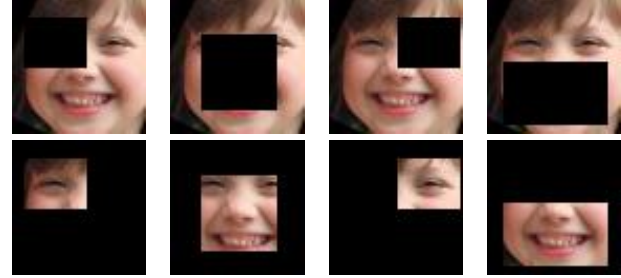}
		\caption{The masked face images are arranged in a sequence from left to right and top to bottom. We refer to the top-row images as occluded images and the bottom-row images as facial component images, for clarity.}
		\label{fig:main_components}
	\end{center}
\end{figure}

It is presented in~\cite{yan2019learning} that covering one of the main components of face images can improve the performance of the kinship network. Hence, we use masked face images for augmentation, such as those without the left or right eye, the nose, or the mouth, as shown in Fig.~\ref{fig:main_components}. For clarity, we call the top-row images occluded images, and the bottom-row images facial component images in Fig.~\ref{fig:main_components}.

Eventually, we employ all 12 augmented images, 4 masked images, and the original image for training CNNs. All images of every identity are randomly split into 14 and 3 for training and testing, respectively. The Adam optimizer with a learning rate of 0.00001 is used. The batch size and the number of epochs are set to 16 and 200, respectively.

The same procedure described in this section is used to train CNNs on high-resolution images generated by GFP-GAN~\cite{wang2021towards}.

\subsection{Configuration of Forest Neural Network(FNN)}
We follow the predefined configuration explained in~\cite{lu2013neighborhood} to have a fair comparison with other papers. Thus, every experiment for every relationship is conducted in the five-fold cross-validation with specified kin or Non-kin relations. Every fold consists of equal Kin (positive) or Non-Kin(Negative) samples. In every iteration, one fold is a test set, and the other folds are a train set. Moreover, at the beginning of each iteration, the model is set to its initial values to avoid affecting other iterations' results. 
The model's performance is the average of five accuracy scores across test sets for each kinship.
The proposed method is implemented in PyTorch 2.3.0 with CUDA 12.1, cuDNN 8.0, and DGL 2.0.0.cu121. The learning rate, batch size, number of epochs, and learning decay are 0.00001, 64, 70, and 0.5, respectively.
The experiments are conducted on a system with an Intel(R) Core(TM) i7-2600k CPU @ 3.40GHz, 12 GB memory, NVIDIA GeForce GTX TITAN X, and Anaconda 24.5.0.

\subsection{Comparison with State-of-the-Art Methods}
We mentioned the experimental results of our proposed method in Table~\ref{table:comparison-to-sota} for the KinfaceW-I and II datasets. Our method achieved the highest results on KinFaceW-II and the highest for the mother-daughter relationship on KinFaceW-I.

\begin{table*}[]
	\caption{Comparison of the cutting-edge algorithms' accuracy with our proposed algorithm for KinFaceW-I and II}
	\centering
	\resizebox{\textwidth}{!}{
		\begin{tabular}{l|ccccc|ccccc}
			Dataset & \multicolumn{5}{c|}{KinFaceW-I}      & \multicolumn{5}{c}{KinfaceW-II}  \\ \toprule
			Method  & F-S   & F-D   & M-S  & M-D   & Mean  & F-S  & F-D  & M-S  & M-D  & Mean \\
			KML~\cite{zhou2019learning}      & 83.8  & 81    & 81.2 & 85    & 82.8  & 87.4 & 83.6 & 86.2 & 85.6 & 85.7 \\
			FSP~\cite{dawson2018same}        & 74.6  & 74.9  & 78.3 & 86    & 76.8  & 92.3 & 84.5 & 80.3 & 94.8 & 90.2 \\
			DDMML~\cite{lu2017discriminative}& 86.4  & 79.1  & 81.4 & 87    & 83.5  & 87.4 & 83.8 & 83.2 & 83   & 84.3 \\
			AdvKin~\cite{zhang2020advkin}    & 75.7  & 78.3  & 77.6 & 83.1  & 78.7  & 88.4 & 85.8 & 88   & 89.8 & 88   \\
			FaCoRNet~\cite{su2023kinship}    & 79.9  & \textbf{81.8}  & 83.9 & 80.6  & 81.55 & 88.6 & \textbf{92.2} & 91 & 90 & 90.6 \\
			H-RGN~\cite{li2021reasoning}     & 81.7  & 78.8  & 81.4 & 88.6  & 82.6  & 90.6 & 86.8 & 93   & 96   & 91.6 \\
			DSMM~\cite{li2021meta}           & 81.7  & 76.7  & 82.3 & \textbf{89}    & 82.4  & 92.6 & 89.8 & \textbf{95.8} & 93.6 & 93   \\
			D4ML~\cite{zhu2023distance}      & \textbf{87.5}  & 79.2  & 80.2 & 88.6  
			& 83.9  & 91.2 & 87.8 & 93.2 & 96.6 & 92.2 \\ 
			FUSE~\cite{nader2025efficient} & 85.6 & 80.4 & \textbf{86} & 87.4 & \textbf{84.85} & 91 & 90 & 93 &92.6 & 91.65 \\
			GADA~\cite{zhao2026generative} & 83.6 & 79.6 & 85.1 & 78.6 & 81.8 & 92.5 & 80.9 & 96.1 & 92.8 & 91.1 \\ 
			\midrule
			FNN     & 82.69 & 79.11 & 77.6 & \textbf{88.95} & 82.09 & \textbf{94.2} & 91.6 & 92.8 & \textbf{96.6} & \textbf{93.8}
		\end{tabular}
	}	
	\label{table:comparison-to-sota}
\end{table*}

\subsection{Effect of using various CNN Networks}
To investigate the significance of each CNN, we conducted experiments to display the accuracy of each network and their concatenations as embedding vectors, separately, as demonstrated in Tables~\ref{table:ablation-kinfacewi} and ~\ref{table:ablation-kinfacewii} for the KinFaceW-I and II datasets, respectively. 

\begin{table*}[pos=htbp]
	\caption{Accuracy Results of various CNNs and their Combinations for the KinFaceW-I dataset. The number inside the parentheses denotes the image's one dimension; for example, 64 indicates that the input image size is 64x64.}
	\label{table:ablation-kinfacewi}
	\centering
	\resizebox{\textwidth}{!}{
		\begin{tabular}{l|cccc|c}
			\textbf{Embedding Network(Input Image Size)} & F-S   & F-D   & M-S   & M-D   & \textbf{Accuracy} \\ \toprule
			ResNet(64)                                   & 73.04 & 66.49 & 66.39 & 75.7  & 70.41             \\
			ResNet(1024)                                 & 78.16 & 67.19 & 69.14 & 77.7  & 73.12             \\
			ResNet(64)+ResNet(1024)                      & 77.91 & 71.31 & 70.63 & 83.77 & 75.91             \\
			ERN(64)                                      & 70.82 & 66.06 & 65.92 & 74.84 & 69.41             \\
			ERN(256)                                     & 71.48 & 64.95 & 70.25 & 79.52 & 71.55             \\
			ResNet(64)+ResNet(1024)+ERN(64)+ERN(256)     & 83.68 & 73.16 & 73.69 & 87.8  & 79.59             \\
			ResNet(64)+ResNet(1024)+ERN(256)             & 81.38 & 76.13 & 77.13 & 88.57 & 80.81             \\ 
			\midrule
			Ensemble of optimal configurations              & 82.69 & 79.11 & 77.6  & 88.95 & 82.09  
		\end{tabular}
	}
	
	\hspace{5mm}
	
	\caption{Optimal Configurations for KinFaceW-I}
	\label{table:kinfacei-best-configurations}
	\begin{tabular}{lll|l}
		$\alpha$ & $H_1$ & $H_2$ & Accuracy \\ 
		\toprule
		1.04     & 256   & 8     & 80.93    \\
		1.05     & 256   & 8     & 80.81    \\
		1.07     & 256   & 8     & 80.58    \\
		1.06     & 256   & 8     & 80.44    \\
		1.03     & 256   & 8     & 80.32   
	\end{tabular}
	
	\hspace{5mm}
	
	\caption{Accuracy Results for the KinFaceW-II dataset.}
	\label{table:ablation-kinfacewii}
	\centering
	\resizebox{\textwidth}{!}{
		\begin{tabular}{l|cccc|c}
			\textbf{Embedding Network(Input Image Size)} & F-S  & F-D  & M-S  & M-D  & \textbf{Accuracy} \\ \toprule
			ResNet(64)                                   & 81.6 & 74   & 78.2 & 82.4 & 79.05             \\
			ResNet(1024)                                 & 88.2 & 82   & 87.4 & 91   & 87.15             \\
			ERN(64)                                      & 81.4 & 76.6 & 78   & 80.6 & 79.15             \\
			ERN(256)                                     & 86.4 & 80.6 & 83.4 & 88   & 84.6              \\
			VGG(64)                                      & 59.6 & 57.8 & 59   & 56   & 58.1              \\
			ResNet(64)+ERN(64)                           & 89   & 84.2 & 86.4 & 92.8 & 88.1              \\
			ResNet(1024)+ERN(256)                        & 90.8 & 88.2 & 91.2 & 94.2 & 91.1              \\
			ResNet(1024)+ERN(256)+VGG(1024)              & 91.8 & 87.6 & 93.8 & 96.2 & 92.35             \\
			ResNet(1024)+ERN(256)+VGG(64)                & 92.4 & 88.6 & 92.8 & 95.6 & 92.35             \\
			ResNet(64)+ERN(64)+VGG(64)                   & 93.8 & 90   & 93   & 96   & 93.2              \\ \midrule
			Ensemble of optimal configurations              & 94.2 & 91.6 & 92.8   & 96.6 & 93.8            
		\end{tabular}
	}
	
	\hspace{10mm}
	
	\caption{Optimal Configurations for KinFaceW-II}
	\label{table:kinfaceii-best-configurations}
	\begin{tabular}{ccc|c}
		$\alpha$ & $H_1$ & $H_2$ & accuracy \\ \toprule
		1.02     & 256   & 8     & 93.3     \\
		1.05     & 256   & 8     & 93.2     \\
		1.05     & 256   & 16    & 93.15    \\
		1.05     & 128   & 8     & 93.05    \\
		1.05     & 256   & 16    & 92.95   
	\end{tabular}
\end{table*}
\begin{table*}[pos=htbp]
	\centering
	\caption{Five runs of random seeds for KinFaceW-II}
	\label{table:kinfaceii-random-runs}
	\resizebox{\textwidth}{!}{
		\begin{tabular}{lccccc}
			Relation& D4ML($\mu\pm\sigma$) & FNN($\mu\pm\sigma)$ & D4ML(95\% CL) & FNN(95\% CL)   & Paired t-test \\ \toprule 
			F-S     & 91.48 ± 0.18        &93.68 ± 0.61         & (91.26, 91.70) & (92.92, 94.44) & t = 6.574, p = 0.0028 \\
			F-D     & 87.00 ± 0.79        &90.00 ± 0.62         & (86.02, 87.98) & (89.23, 90.77) & t = 5.809, p = 0.0044     \\
			M-S     & 91.84 ± 0.33        &92.76 ± 0.48         & (91.43, 92.25) & (92.17, 93.35) & t = 3.683, p = 0.0211    \\
			M-D     & 95.60 ± 0.55        &96.12 ± 0.11         & (94.92, 96.28) & (95.98, 96.26) & t = 2.525, p = 0.0650   \\
		\end{tabular}
	}
\end{table*}

The ResNet architecture is unchanged, except that we applied average pooling to the feature map before feeding it to the FNN. We redesigned the ERN architecture to align its feature dimension with those of the other networks. Hyperparameters are the same as the general optimal values (such as $\alpha=1.05$ and $\omega_0=0.01$).
The parameters of the ensemble for KinFaceW-I and II are mentioned in Table~\ref{table:kinfacei-best-configurations} and Table~\ref{table:kinfaceii-best-configurations}, respectively.

The results indicate that CNNs and their combinations affect accuracy because each serves as an expert, computing based on its own perspective on the data. 

\subsection{Architecture of Kinship Classification}
Initially, we used two fully connected layers without hidden layers. However, when the center loss is incorporated with the binary cross-entropy loss in this network, it did not improve accuracy. Hence, we redesigned the kinship classification module to use center loss and added hidden layers to improve accuracy. The main reason for adding hidden layers is to ensure that the center loss affects the class spread in the feature space.
Additionally, the MLP's input layer contains a large number of neurons, while the output layer comprises only one or two neurons. Hence, the first hidden layer is assumed to be bigger than the second. Nonetheless, the sizes of these hidden layers are hyperparameters that must be examined. The various values of the hidden layers are probed and illustrated in Table~\ref{table:kinship-module} for KinFaceW-II. When the sizes of these layers are (256,8) or (256,16), the mean accuracy of all relationships is the highest. Tables \ref{table:kinfacei-best-configurations} and \ref{table:kinfaceii-best-configurations} presented the top five configurations based on the mean accuracy for KinFaceW-I and II, respectively.

\begin{table}[pos=htbp]
	\captionsetup{width=\linewidth} 
	\caption{Various sizes of the first and second hidden layers.}
	\centering
	\begin{tabular}{l|cccc}
		\diagbox{$H_1$}{$H_2$} & 8     & 16    & 32    & 64    \\ \hline
		128     & 93.05 & 92.95 & 92.8  & 92.5  \\
		192     & 92.4  & 92.45 & 92.7  & 92.35 \\
		256     & \textbf{93.2}  & \textbf{93.15} & 92.55 & 92.3  \\
		512     & 92.4  & 92.5  & 92.15 & 91.8  \\
	\end{tabular}
	\label{table:kinship-module}
\end{table}

\subsection{Hyper-parameter Analysis}
Hyperparameters in our method are the weights used to combine different loss functions, which need to be analyzed and tuned to optimal values, as they can scale losses and weigh their impacts on the model.

The weights $\omega_i$ match those of D4ML~\cite{zhu2023distance} and exhibit the same functionality and performance in our work.
We also have two hyperparameters, meters $\omega_0$ and $\alpha$ related to the center loss. Our optimization approach is to fix all other hyperparameters except the one we want to find. The reported accuracy is the mean across all kinship types.

\begin{figure*}[pos=htbp]
	\begin{center}
		\includegraphics[width=0.45\textwidth]{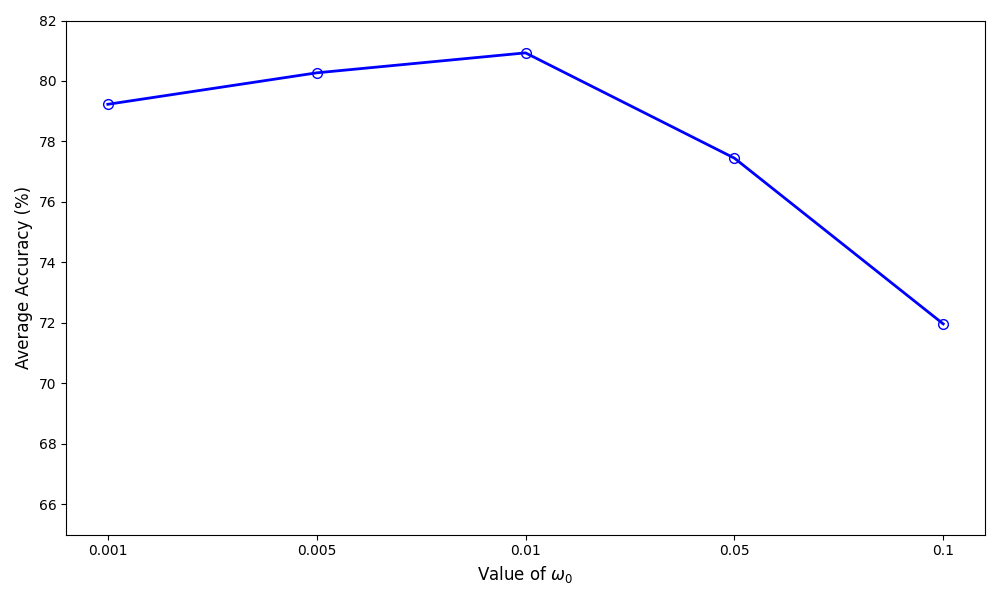}
		\hspace{0.05\textwidth}
		\includegraphics[width=0.45\textwidth]{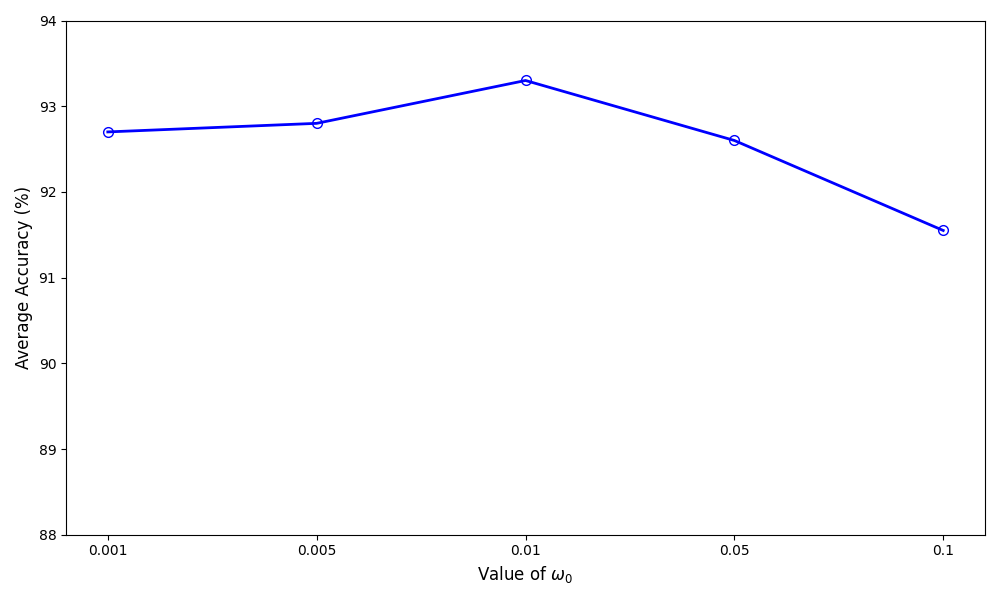}
		\caption{The initial temperature for KinFaceW-I and II. The effects of different initial temperatures were investigated while other parameters were held constant.}
		\label{fig:initial_temperature}
	\end{center}
\end{figure*}

\begin{itemize}
	\item \textit{Impact of} $\omega_0$:
\end{itemize}
The figure~\ref{fig:initial_temperature} shows that we examined the various values of the initial parameter over [0.001, 0.1], and 0.01 was the optimal value, as other values degrade accuracy.

\begin{figure*}[pos=htbp]
	\begin{center}
		\includegraphics[width=0.45\textwidth]{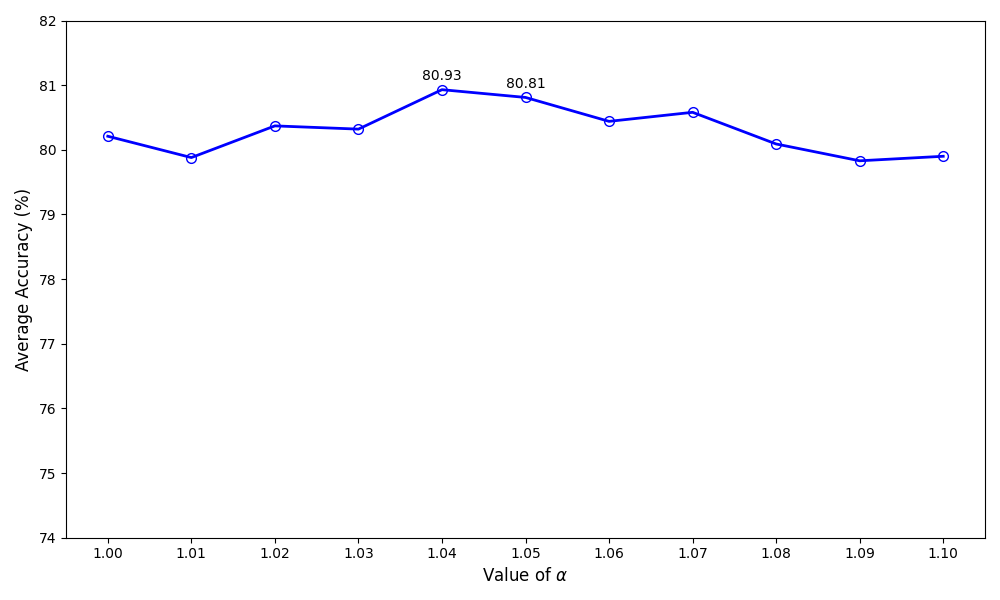}
		\hspace{0.05\textwidth} 
		\includegraphics[width=0.45\textwidth]{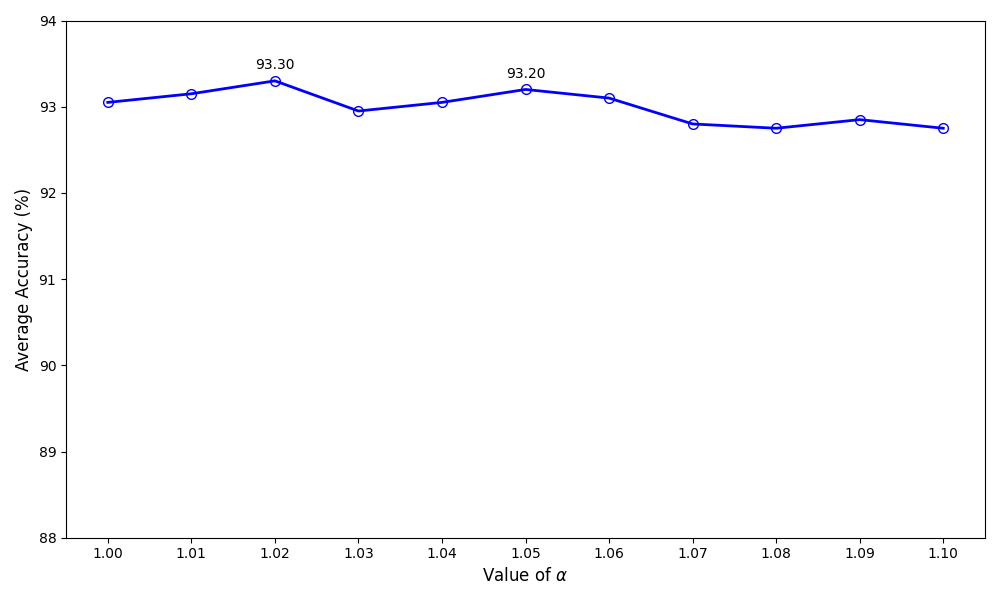}
		\caption{The Temperature rate $\alpha$ for KinFaceW-I and KinFaceW-II. The optimal values for KinFaceW-I were 1.04 and 1.05, and for KinFaceW-II were 1.02 and 1.05.}
		\label{fig:temperature_rate}
	\end{center}
\end{figure*}

\begin{itemize}
	\item \textit{Impact of} $\alpha$:
\end{itemize}
The search range of the optimal value for the temperature rate $\alpha$ is [1,~1.1] for KinFaceW-I and KinFaceW-II, represented in Fig.~\ref{fig:temperature_rate}. The optimal values for KinFaceW-I were 1.04 and 1.05, and for KinFaceW-II were 1.02 and 1.05. We selected 1.05 for the temperature rate because it was a common value that achieved high accuracy across both datasets. Values less than one do not make sense because they gradually diminish the impact of the center loss. Values greater than 1.10 noticeably degrade accuracy and are not depicted in Fig.~\ref{fig:temperature_rate}.

\subsection{Statistical Validation of Results on KinFaceW-II}
To check the results are reliable and repeatable, we conducted a statistical validation of experiments using D4ML as a baseline method and our proposed method, FNN, on the KinFaceW-II dataset. 
Each of them was trained and evaluated by randomly and independently setting seeds five times. We collected results and reported the sample mean, sample standard deviation, and a 95\% confidence level for each relationship type on KinFaceW-II. 
Moreover, we performed paired t-tests to check whether our proposed method has statistically significant improvements over D4ML, as shown in Table~\ref{table:kinfaceii-random-runs}. It revealed that our proposed method showed statistically significant improvements in the F-S and F-D relationships (p$<$0.01) and in the M-S relationship (p$<$0.05). For M-D, it indicated that the intervals slightly overlaped, and the improvement was minor.

\subsection{Effect of the Center Loss}
It is essential to assess whether center loss affects the accuracy of kinship verification. If its presence does not raise the accuracy, it does not need to be included in the combination of losses. Therefore, we conducted experiments to investigate its presence in the loss fusion. 

\begin{figure*}[pos=htbp]
	\begin{center}
		\includegraphics[width=0.45\textwidth]{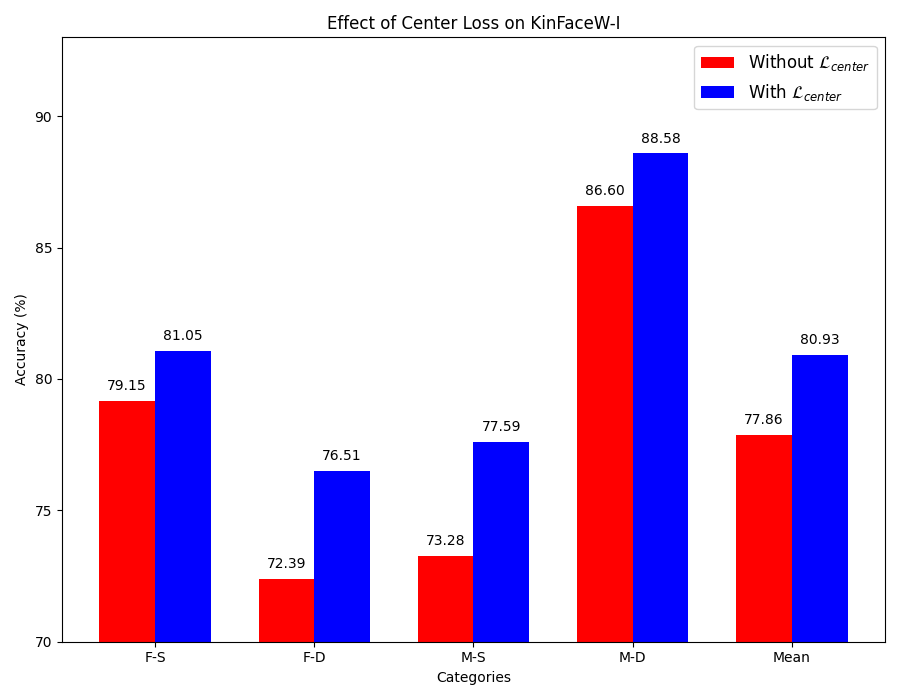}
		\hspace{0.05\textwidth} 
		\includegraphics[width=0.45\textwidth]{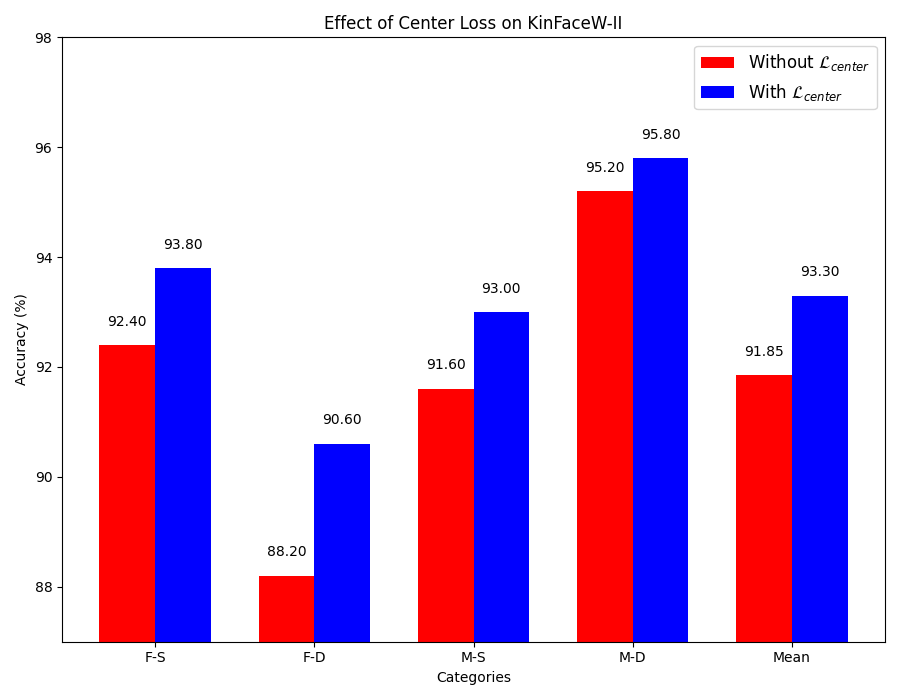}
		\caption{Results of our proposed method with and without using the center loss on the KinFaceW-I and KinFaceW-II datasets.}
		\label{fig:centerloss_effect}
	\end{center}
\end{figure*}

We demonstrated the results of the KinFaceW-I and KinFaceW-II for the presence or absence of the center loss in the fusion of losses in Fig.~\ref{fig:centerloss_effect}.
Its presence can improve accuracy in all kinship types. For example, it improved the accuracy of FS:1.9, FD:4.12, MS:4.31, and MD:1.98, with a mean improvement across all types of 3.07. It achieved an accuracy of at least 2 percent, and generally 3 percent, on the KinFaceW-I dataset. In addition, it enhanced the accuracy of FS:1.4, FD:2.4, MS:1.4, and MD:0.6, and the overall mean accuracy across all types is 1.45 on the KinFaceW-II dataset.

Therefore, adding the center loss can improve accuracy by nearly 2 percent when combined correctly with other losses. Also, its presence in the KinFaceW-I is more substantial, indicating that it can help the network better separate two classes when we have diverse image sources.

\subsection{Effect of Each Loss}
\begin{figure}[pos=htbp]
	\begin{center}
		\includegraphics[width=\textwidth]{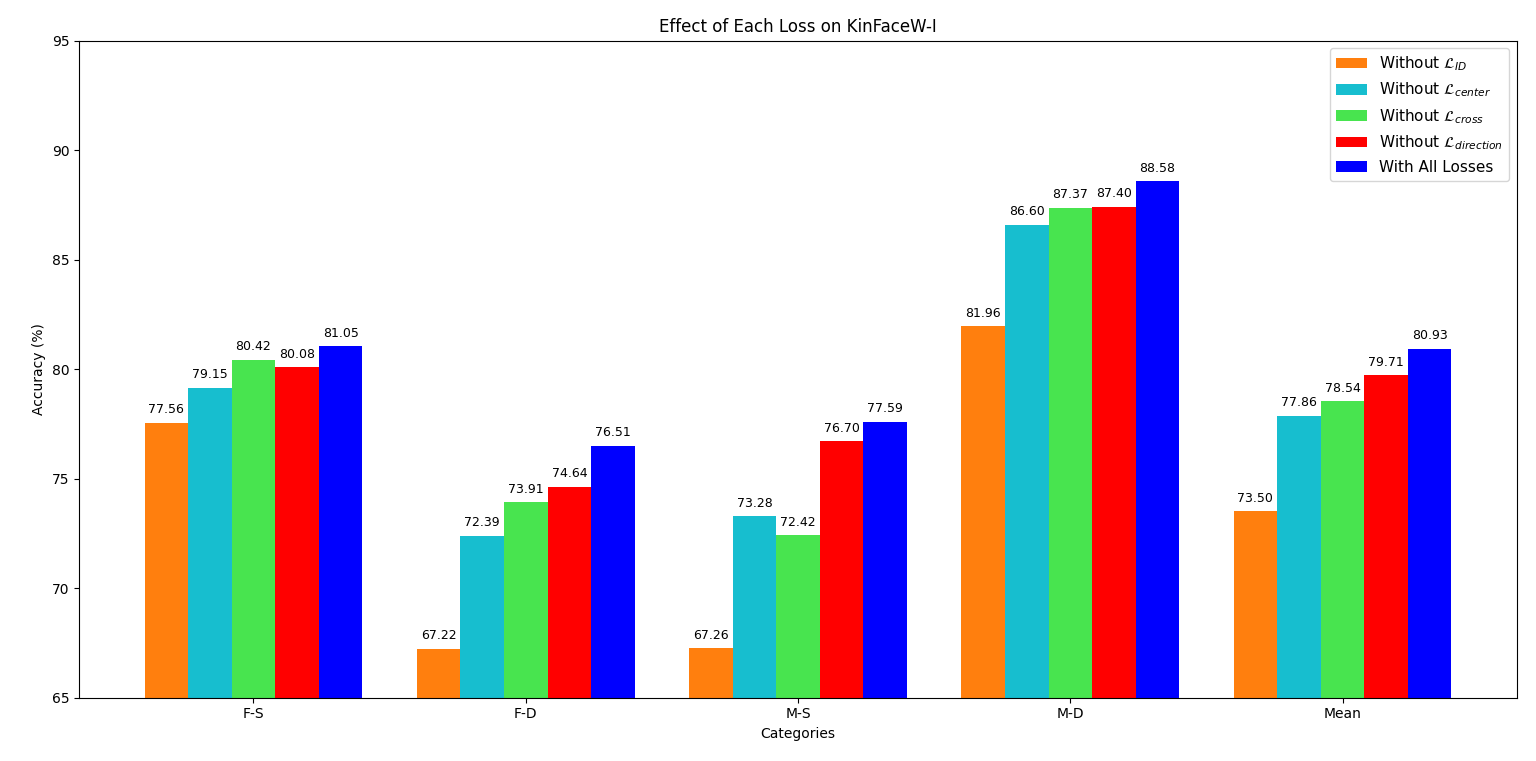}
		\hspace{0.05\textwidth} 
		\includegraphics[width=\textwidth]{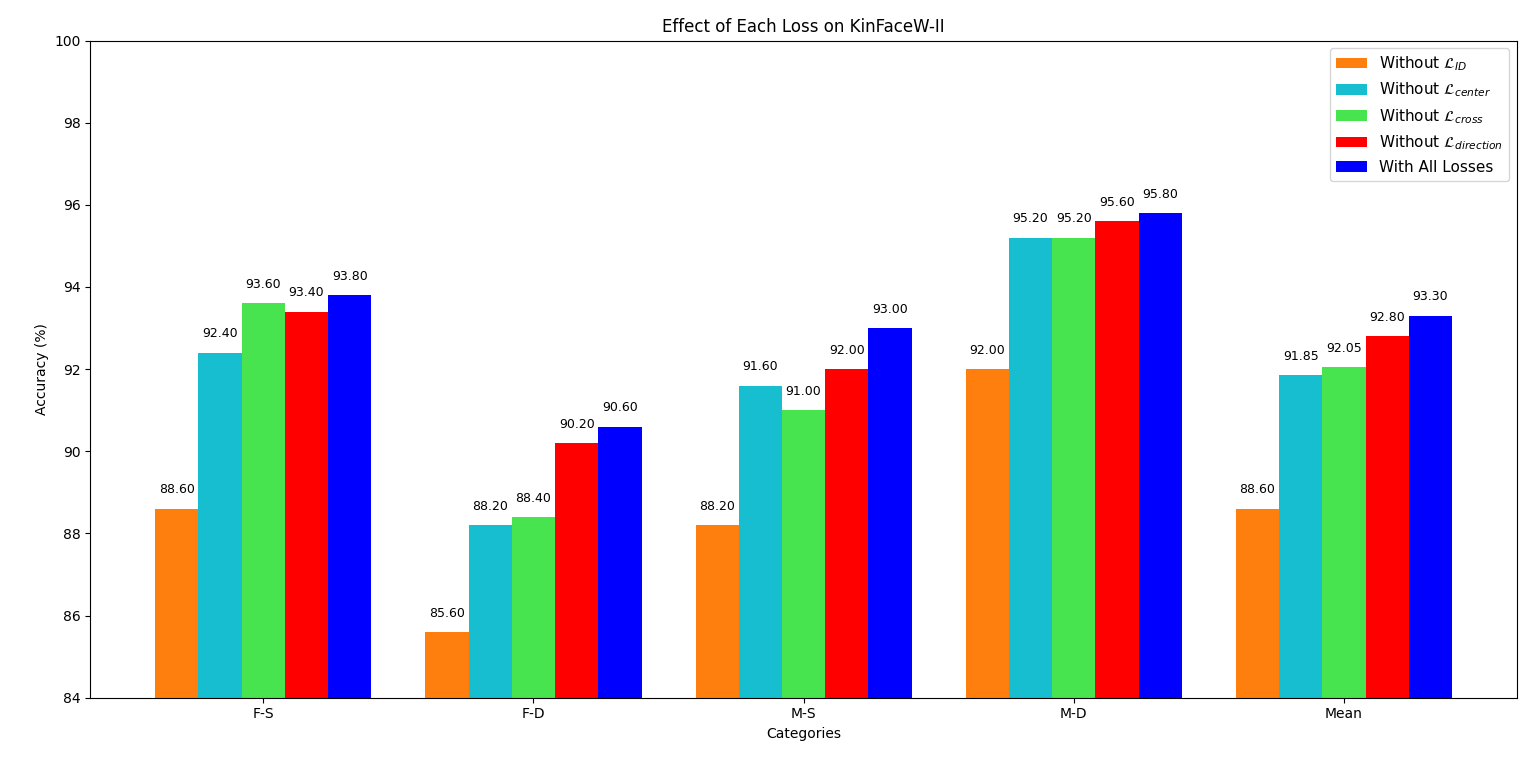}
		\caption{Results of our proposed method with and without using different losses on the KinFaceW-I and KinFaceW-II datasets.}
		\label{fig:eachloss_effect}
	\end{center}
\end{figure}

It is vital to investigate the contribution of each loss to improving accuracy. Therefore, we conducted experiments on all four relationships across both datasets, KinFaceW-I and II, and presented the results in  Fig.~\ref{fig:eachloss_effect}. When we reported results without loss ($L_{i}$), it means that we performed the experiments with all other losses except loss ($L_{i}$).

Furthermore, we added additional experiments to assess results across kinship types, and reported them in Fig.~\ref{fig:eachloss_effect}. 
The results generally indicate the priority of various losses in our algorithm: the losses $\mathcal{L}_{id}$, $\mathcal{L}_{center}$, $\mathcal{L}_{cross}$, and $\mathcal{L}_{direction}$ are essential in order.

\subsection{Effect of High-resolution Images}
We used GFP-GAN~\cite{wang2021towards} to generate high-resolution images and applied the same procedures to the original photos as to the high-resolution images. After training, the test results were reported in Table~\ref{table:ablation-kinfacewi} and Table~\ref{table:ablation-kinfacewii}. ResNet with 1024×1024 input images achieved 2.64\% and 8.1\% higher accuracy than with 64×64 images, on average across the four kin types, on KinFaceW-I and II, respectively. 
We obtained the same results for 256×256 high-resolution photos. Therefore, to reduce computation, we trained the ERN network only on 256×256 images and achieved 5.45\% higher average accuracy than with 64×64 images across the four kin types on KinFaceW-II. 
For KinFaceW-I, we obtained improvements in FS (0.66\%), MS (4.33\%), and MD (4.68\%). However, it did not increase the FD relation; it decreased by 1.11\%. 

In conclusion, using high-resolution images generally improves the results on both datasets. 


\subsection{ROC Analysis}
The ROC curve illustrates a model's diagnostic performance. We can control the number of false positives and set a threshold based on the accuracy we need in our application, especially in surveillance systems and sensitive applications such as forensic analysis. 
Its axes are the true positive rate (TPR) and the false positive rate (FPR). The true positive rate is the proportion of correctly predicted positive relations among all real positive relations (kin-related samples). 
On the other hand, the false positive rate is the proportion of incorrectly predicted positive relations among all real negative relations (non-kin samples) as stated in Eq.~\ref{eq:tpr}.  

\begin{equation}
	TPR=\frac{TP}{TP+FN},~~~ FPR=\frac{FP}{TN+FP}
	~\label{eq:tpr}
\end{equation}

\begin{figure}[pos=htbp]
	\begin{center}
		\includegraphics[width=0.45\textwidth]{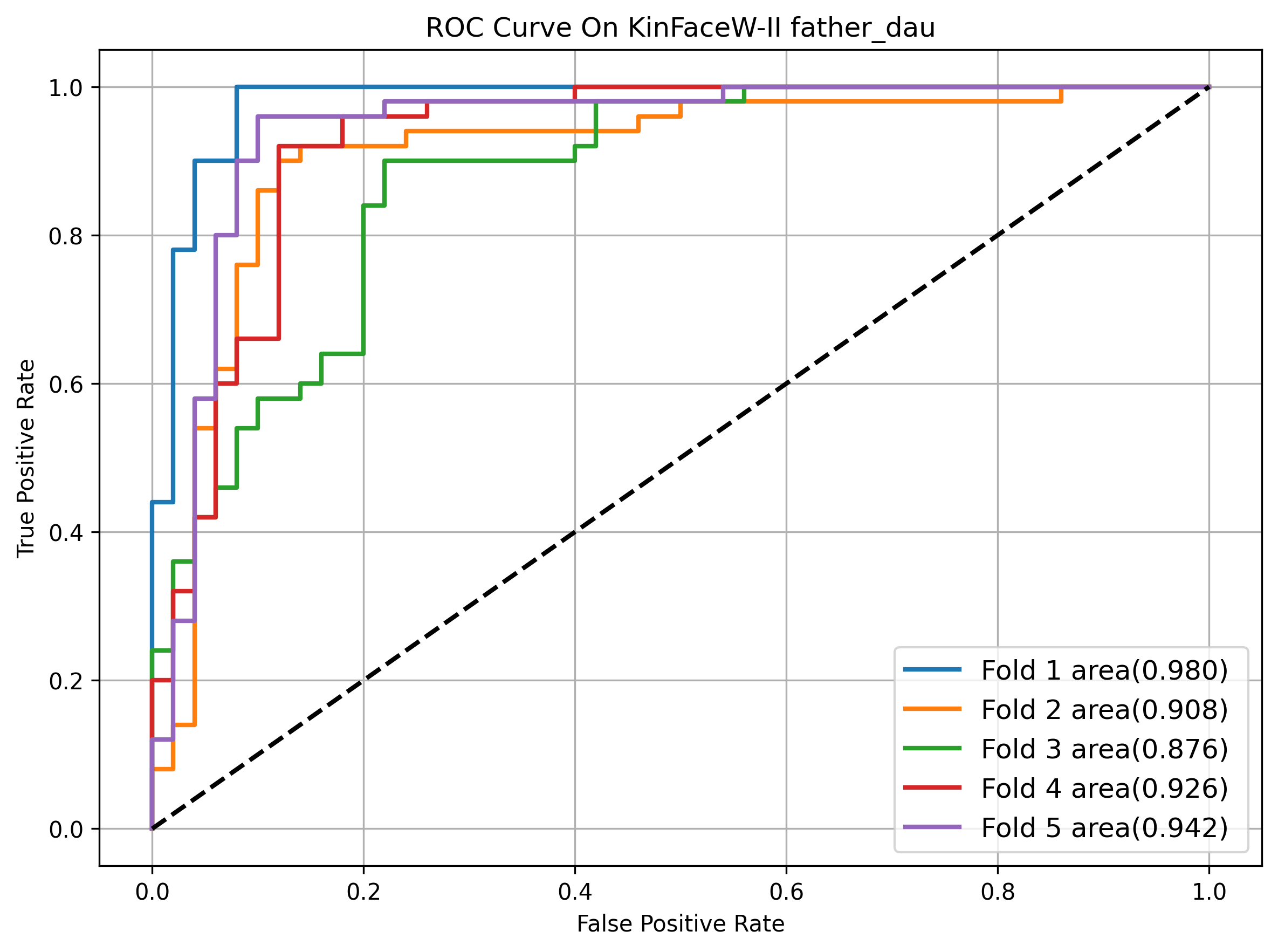}
		\hspace{0.05\textwidth} 
		\includegraphics[width=0.45\textwidth]{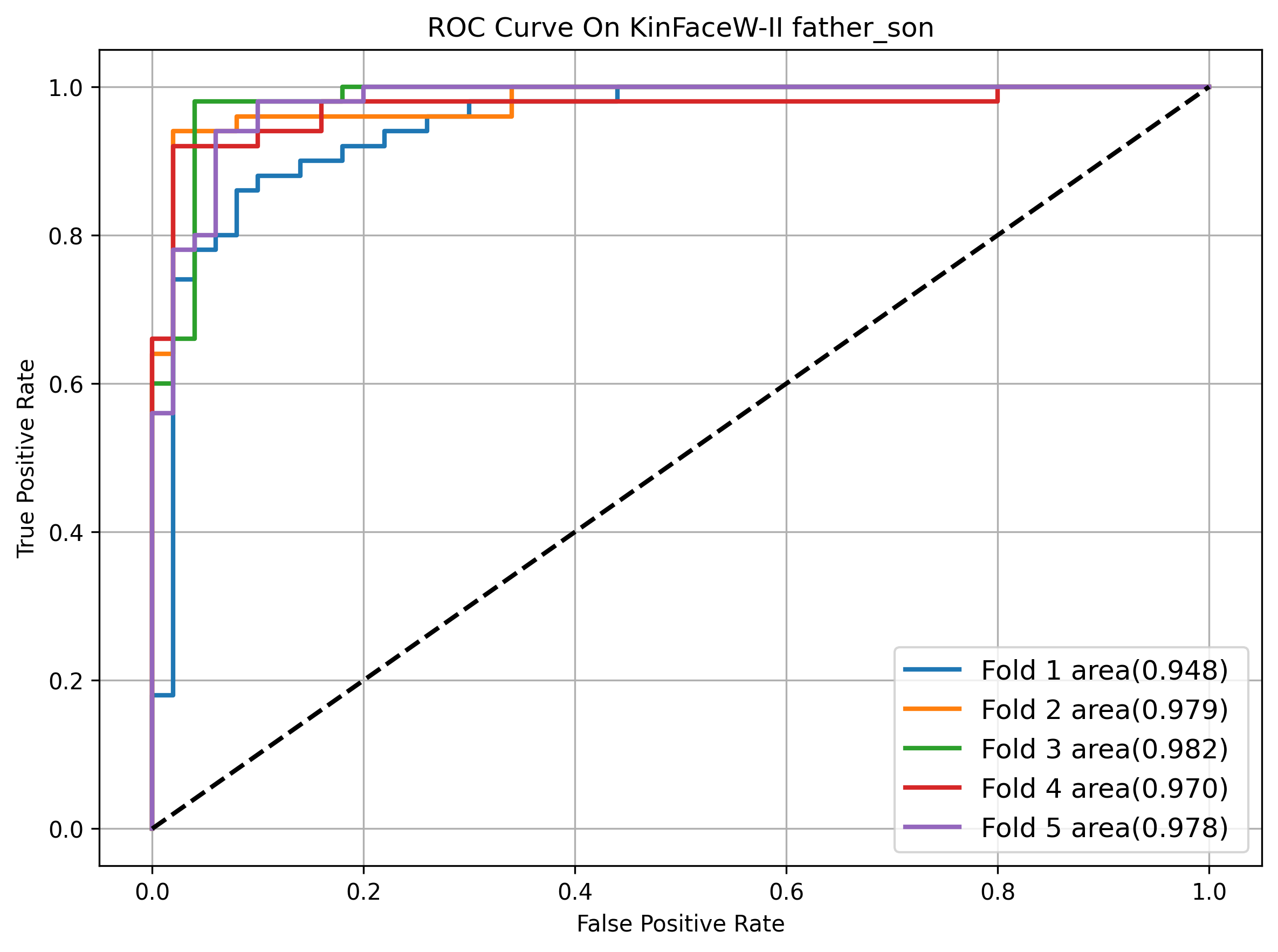}
		\includegraphics[width=0.45\textwidth]{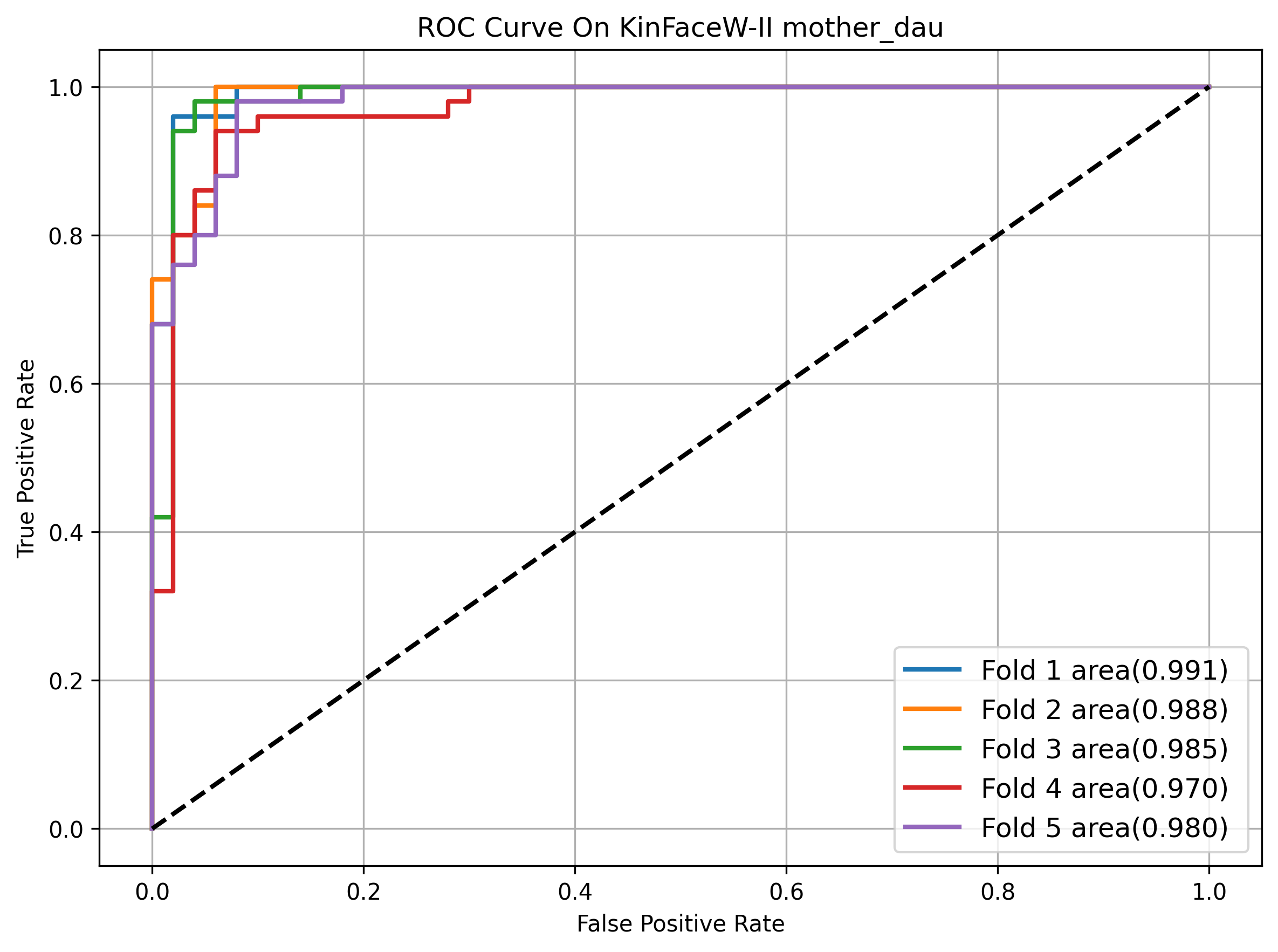}
		\hspace{0.05\textwidth} 
		\includegraphics[width=0.45\textwidth]{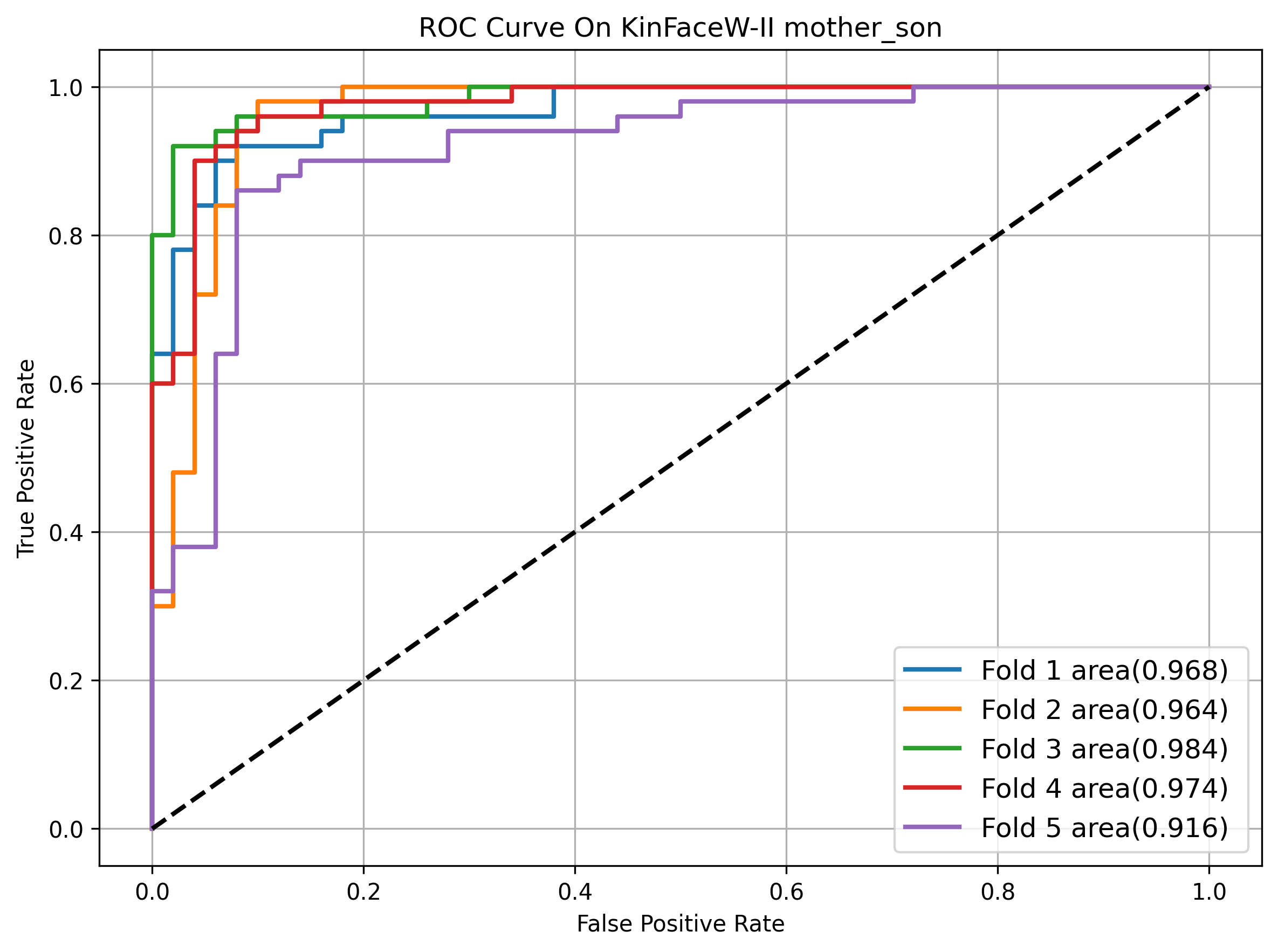}
		\caption{The ROC curves of the four relationships (FD, FS, MD, and MS) for KinFaceW-II.}
		\label{fig:roc-kinfacewii}
	\end{center}
\end{figure}

We drew ROC curves with their AUCs (Area Under Curve) for four relationships of KinFaceW-II shown in Fig.~\ref{fig:roc-kinfacewii}. We can also observe that fold 3 of the F-D relation was the hardest and had the lowest accuracy.

\subsection{Computational Complexity and Model Size Analysis}
\label{subsec:flops}
To compute the cost of our proposed method with D4ML and avoid dependence on specific hardware, we measure the number of floating-point operations (FLOPs) and trainable parameters using the \texttt{fvcore}~\cite{fvcore} tool and the model size using \texttt{torchinfo}~\cite{torchinfo} tool.

For our graph-based model, FNN, we use nine graphs with two nodes and two edges. The feature dimensions of every node and edge are 2688 and one for KinFaceW-II, respectively. These dimensions are 1536 and one for KinFaceW-I, respectively.
For D4ML, a pair of image inputs of size 3 × 64 × 64 is used. 
We also compared our model size with the D4ML algorithm and showed that our proposed model is much smaller and has fewer training parameters. We computed training FLOPs by summing the FLOPs of the forward pass, backward pass, and model update.
We provided the results of these comparisons in Table~\ref{table:flops}.

FLOPs analysis revealed that, since D4ML used one CNN and fully connected layers, its FLOPs were high, on the order of $10^{9}$. In contrast, FNN achieved a reduction in FLOPs of approximately 1.72, 2.31, 2.39, and 2.22 times for KinFaceW-II, and a reduction of 5, 6.3, 6.3, 6.35 times for KinFaceW-I in total parameters, forward FLOPs, training FLOPS per iteration, and size by using facial image embeddings.

Regarding the computation, the results revealed the advantage of an FNN over D4ML when facial image embeddings are available. 

\begin{table}[pos=htbp]
	\caption{Comparison of computational complexity between \texttt{D4ML} and \texttt{FNN}.}
	\begin{tabular}{lllll}
		Model    & Total Parameters & Total Forward FLOPs & \begin{tabular}[c]{@{}l@{}}Training FLOPs\\ per Iteration\end{tabular} & size(MB) \\ \toprule
		D4ML  & 91.33 M & 230,181,120   & 690,543,360   & 462.78 \\ \midrule
		FNN(KinFaceWII) & 53 M  & 99,373,176 & 289,119,528 & 208.82   \\
		FNN(KinFaceWI) &   18.3 M  & 36,520,976    & 109,562,928   & 72.83  \\
	\end{tabular}
	\label{table:flops}
\end{table}

\subsection{Performance Visualization}
\label{subsection:performance-visualiztion}
To demonstrate the behaviour of our algorithm, we plotted the various losses and test accuracy of the F-D relation of KinFaceW-II over 50 epochs in Fig.~\ref{fig:visualization-center-loss} and Fig.~\ref{fig:visualization-test-accuracy}, respectively. Fig.~\ref{fig:visualization-center-loss} highlights the influence of the center loss during training in conjunction with other losses and its presence when another loss is absent.
Furthermore, we plotted test accuracy from five-fold cross-validation, using the folds as test sets, in Fig.~\ref{fig:visualization-test-accuracy} over 50 epochs.

\subsection{Impact of each Facial Component and its Occlusion}
\label{subsection:impact-facial-component-occlusion}
We conducted experiments on KinFaceW-II to investigate the influence of each facial region and its occlusion on our proposed algorithm. We reported the results in Table~\ref{table:effect-each-component}, which shows that when one facial component or its occlusion from both the parent and child is not on the graph, the KV accuracy remains near the highest. In addition, the nose and eyes, in order, are essential for preserving accuracy and align with cognitive science findings that the nose and eyes are significant in identifying people and kinship. We also observed that when the eyes are occluded, but the nose and mouth are not, the KV accuracy does not drop much. 

In conclusion, our method is robust to the occlusion of a single facial component. Additionally, the nose, eyes, and mouth regions are significant in order, and their presence alongside the features of the entire face increases the KV accuracy.

\section{Discussion and Analysis}

\subsection{Comparison with Graph-Based KV Methods}
\label{subsection:comparison-graph-based}

Weighted Graph Embedding-based Metric Learning (WGEML)~\cite{liang2018weighted} employed a traditional metric learning method to address the KV problem by combining hand-crafted descriptors, such as LBP, HOG, SIFT, and CNN (VGG-Face). For each feature type, one intrinsic graph was defined to pull the positive pairs close, and two penalty graphs — one to push away negative pairs and one for hard negatives — were defined. Scatter matrices were constructed from these graphs to form an objective function and to identify an optimal metric for each feature type. Learned metrics were combined based on their weight coefficients. 

One of its major drawbacks is that it is poorly suited to large datasets, as it relies on matrix decomposition and iterative optimization. Our proposed method differs significantly from WGEML: it uses graphs to construct scatter matrices and optimize the objective function, whereas we perform all computations and updates on graphs. WGEML optimized the objective functions iteratively, whereas our method uses a neural network and optimizes via gradient descent.

Additionally, Reasoning Graph Networks~\cite{li2021reasoning} proposed two graph structures: the Star-shaped Reasoning Graph Network (S-RGN) and the Hierarchical Reasoning Graph Network (H-RGN). S-RGN constructed a star-shaped graph by comparing each dimension of the parent and child feature vectors. A central node, connected to all comparison nodes, received the information from all compared nodes. The drawback of this method is that having only one central node for aggregating comparisons of parent-child information is insufficient and yields lower accuracy. 

Their authors proposed another structure, a hierarchical tree-like structure, H-RGN, to aggregate comparison information from lower to higher layers. Then, the aggregated information from the top flowed to the bottom. 

A comparison of our graph structure with others reveals that ours differs completely. In the forest, we connect only two nodes, and each node has its own features. We do not compare the dimensions of the parent and child feature vectors; instead, we exchange the entire information of the parent and child nodes. 

Both S-RGN and H-RGN processed feature vectors either one-by-one or divided them into parts, without considering face components or whole feature vectors. Furthermore, the aggregations of S-RGN and H-RGN were entirely different from ours: they concatenated all nodes and sent them to an MLP, whereas we used the mean aggregation across all nodes. In addition, they did not consider various feature types and losses.

\subsection{Comparison with Facial Component-Based Methods}
\label{subsection:comparison-facial-component-based}
FaCoRNet~\cite{su2023kinship} considered the idea of the facial components’ resemblance between parent and child. However, it used only the feature maps of parent-child facial images and a cross-attention mechanism to verify kinship. It did not extract facial components; instead, it explored them in the feature maps. In addition, it used a marginalized contrastive loss to pull kin pairs closer and push non-kin pairs further apart, whereas we did not use this loss in our algorithm. 

Regarding the accuracy comparison, we included this method alongside others in Table~\ref{table:comparison-to-sota}. The KV accuracy of our FNN is 0.54 and 3.2 percent on average across all kin types, higher than FaCoRNet for KinFaceW-I and KinFaceW-II, respectively. FaCoRNet achieved higher accuracy than FNN for the F-D relationship on both datasets, and higher accuracy for M-S on KinFaceW-I. Whereas in the other relationships, our FNN achieved higher accuracy.

\section{Conclusion}
\label{sec:conclusion}
We introduced an innovative algorithm, Forest Neural Network~(FNN), that leveraged face classification algorithms for kinship verification by exploiting their face representations and incorporating graph neural network operators. Our proposal outlined the application of each CNN network as an expert system and fed its outputs into FNN. This graph-based algorithm accounted for the disjoint nature of facial components, fused their processed features to enhance kinship verification accuracy, and achieved performance on par with the cutting edge. Results provided reassurance about the effectiveness of our approach in kinship verification.

Moreover, we proposed a new combination of loss functions, particularly including the center loss, to gradually train our architecture and the kinship classification module. The outputs of the fully connected layers were given to the center loss to compress the features of each class and separate classes from each other.

In addition, we demonstrated the effectiveness of high-resolution images on KinfaceW-I, as photos were collected from various sources. Converting them to high resolution can mitigate the differences among images and enhance classification accuracy.

Furthermore, the proposed algorithm can be used alongside a face recognition system, eliminating the need to train kinship verification systems from raw facial images.

\section{Future Work}
One promising approach is to explore various combinations of new popular networks to improve the accuracy of kinship verification, such as EfficientNet and ViTs. 
Furthermore, investigating hyperparameters and model parameters for end-to-end learning of the proposed architecture may lead to significant advancements in the field. 

Moreover, the proposed algorithm can be applied to other kinds of kinship datasets, such as TSKinFace~\cite{qin2015tri} and the large-scale kinship dataset FIW~\cite{robinson2018visual}, to evaluate its performance on them. Furthermore, because we followed the standard configuration of KinFaceW-I and KinFaceW-II, we were not authorized to investigate demographic biases such as race. This development may be a promising avenue for research in building fair models. 

Moreover, another promising avenue is to develop reliable methods to handle false-positive samples in sensitive applications, such as forensic analysis. 
In addition, the subjects of demographic bias, misuse in surveillance, and ethical concerns are essential, especially in high-stakes operational settings, and are necessary for conducting extensive research. 
Another scenario to extend the proposed algorithm is to occlude facial components during the evaluation phase to assess its performance. Since we occluded one component of faces during training, it is more reasonable to expect greater robustness than the others. 

Furthermore, another way to enhance FNN is to analyze its failure cases for understanding the model's behavior and determine whether there is a failure pattern across the four types. 

Investigating failure patterns may reveal significant demographic differences, such as large age gaps, variation in facial expressions, and occlusions. Additionally, this failure stems from specific characteristics of the relationship.

\appendix

\section{Figures of Performance Visualization}
\label{appendix:performance-visualization}

We investigated the contribution of the center loss and demonstrated its positive role in the loss function and test accuracy in Fig.~\ref{fig:visualization-center-loss} and~\ref{fig:visualization-test-accuracy}.

\subsection{Figures of Effect of the Center Loss}
\label{appendix:performance-visualization-center-loss}

\begin{figure}[pos=htbp]
	\begin{center}
		\includegraphics[width=0.45\textwidth]{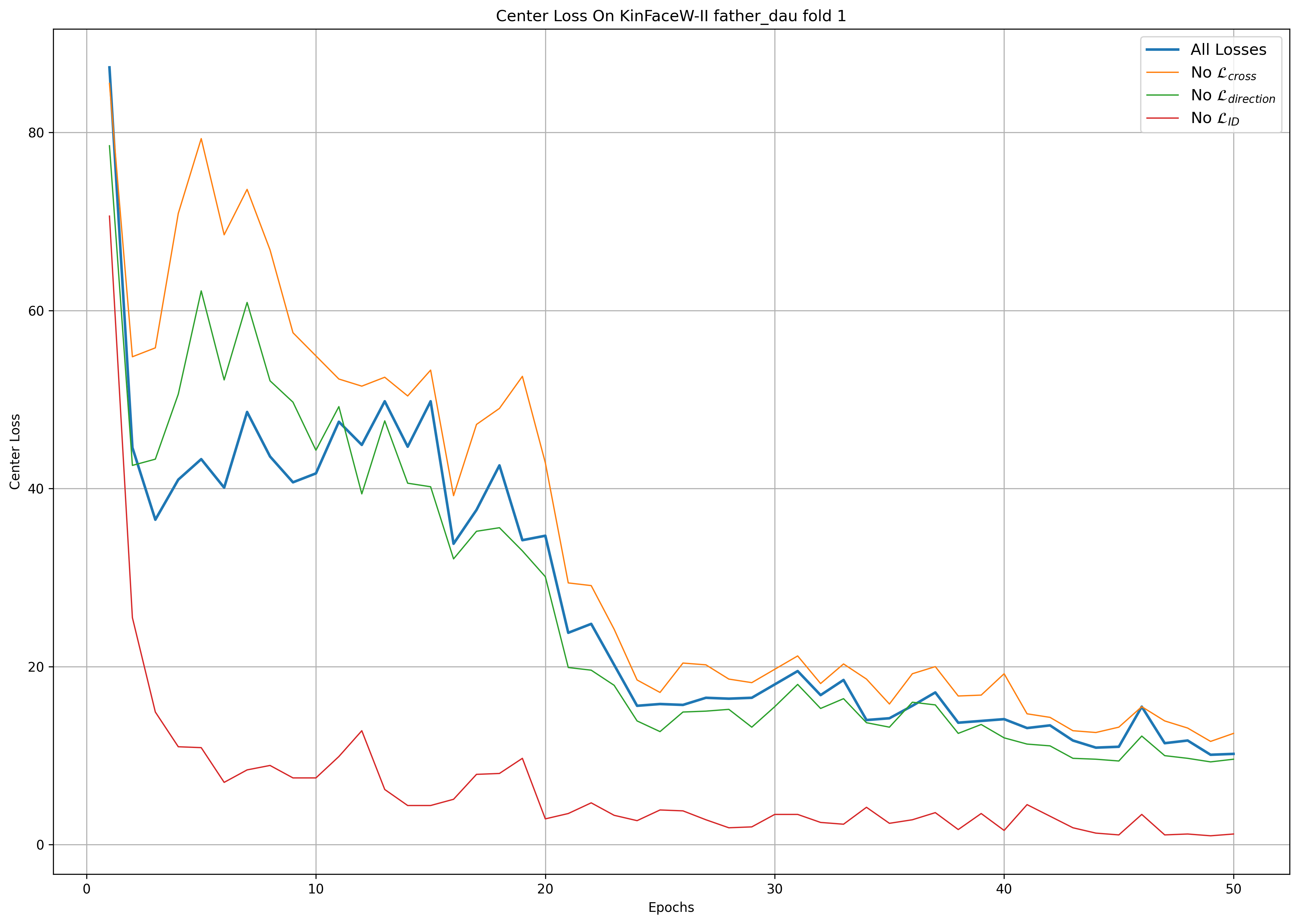}
		\hspace{0.05\textwidth} 
		\includegraphics[width=0.45\textwidth]{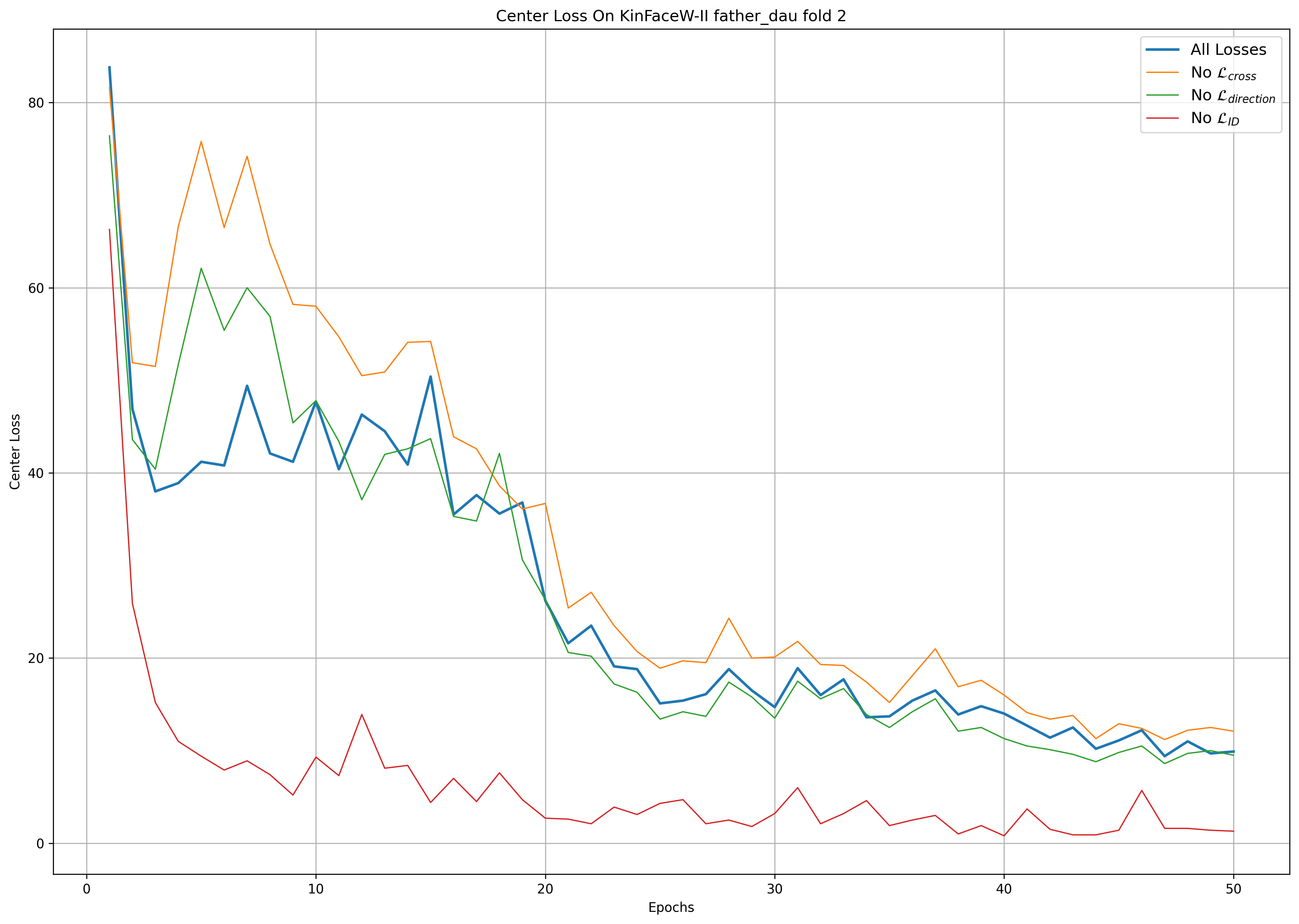}
		\hspace{0.05\textwidth} 
		\includegraphics[width=0.45\textwidth]{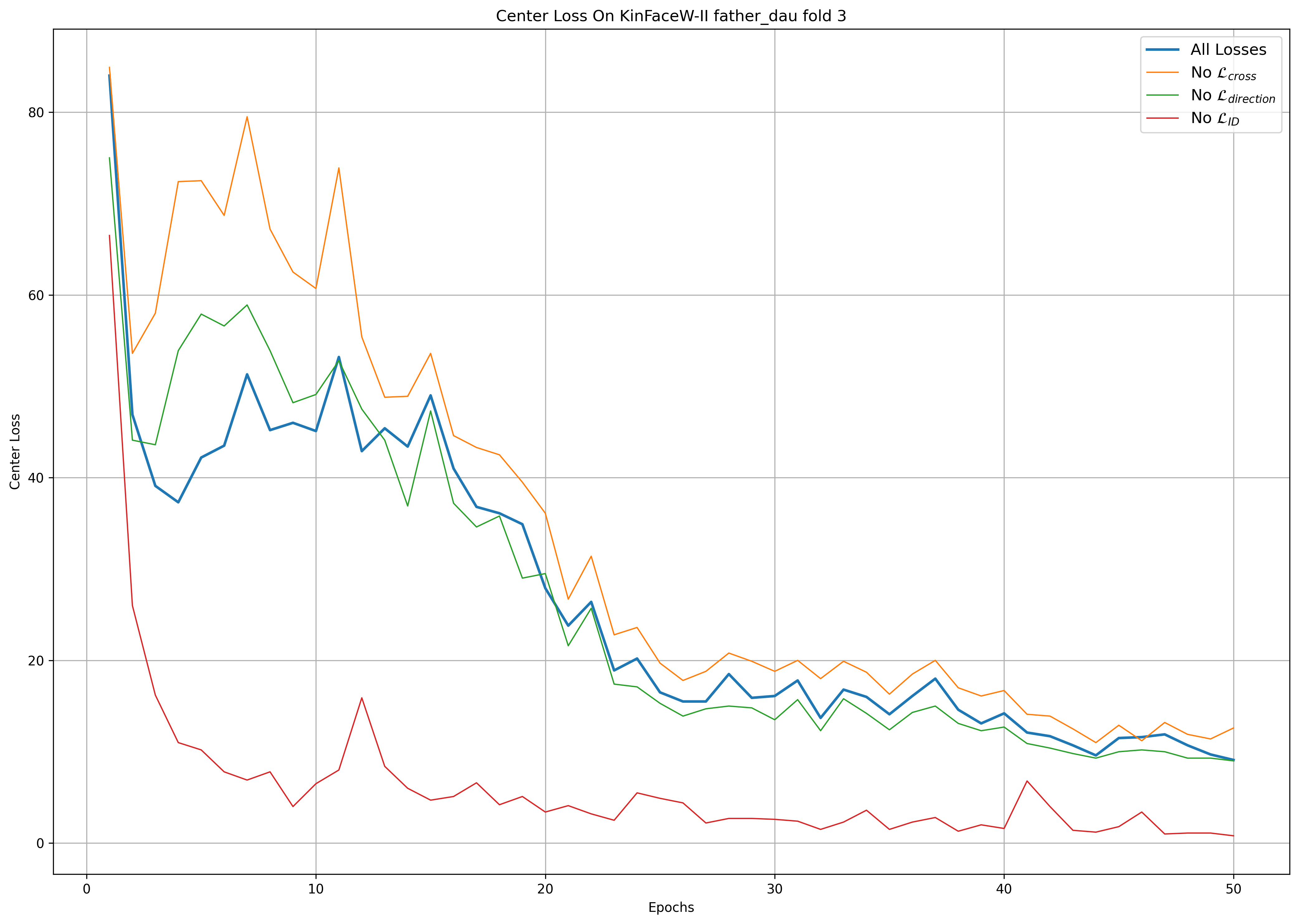}
		\hspace{0.05\textwidth} 
		\includegraphics[width=0.45\textwidth]{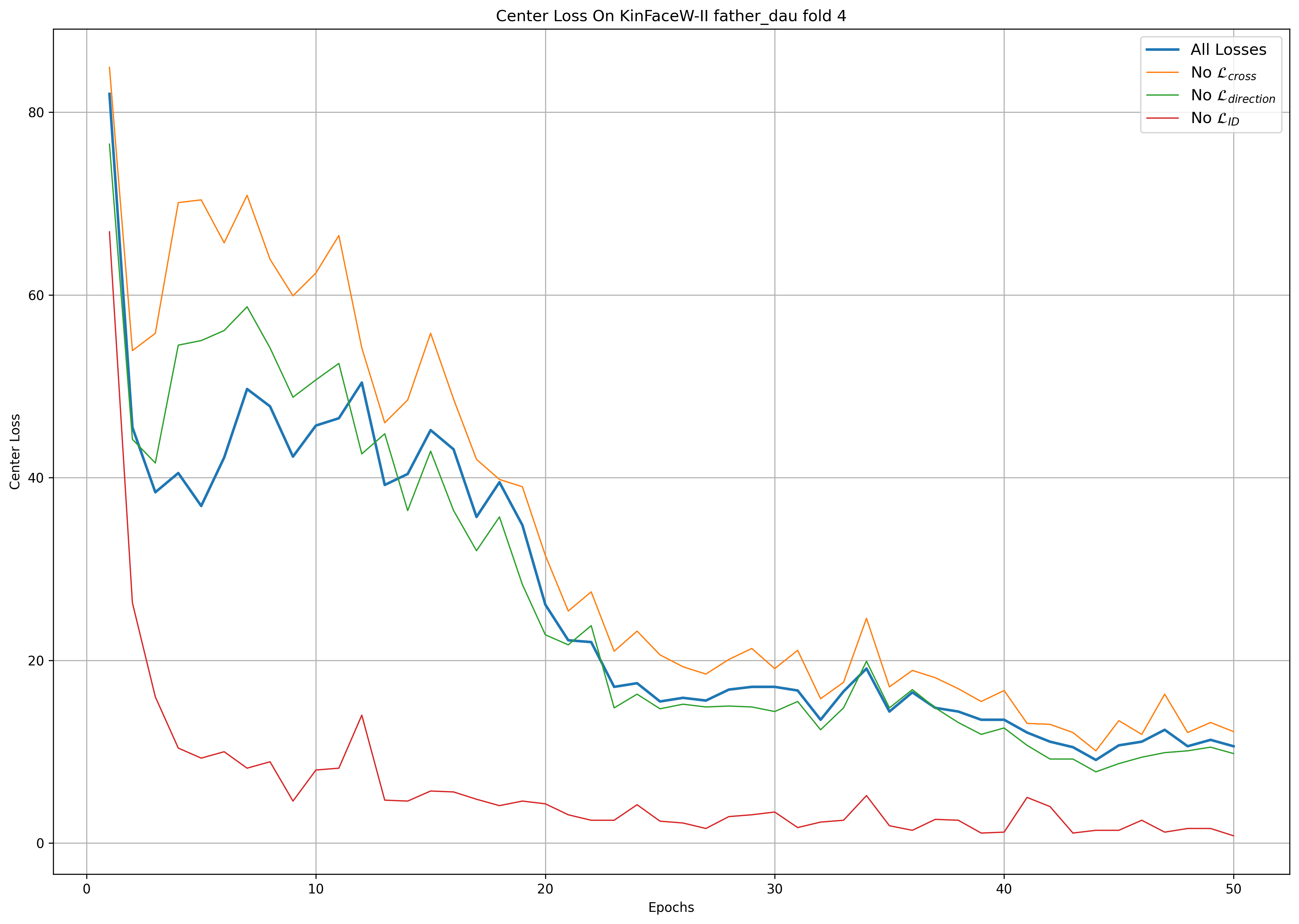}
		\hspace{0.05\textwidth} 
		\includegraphics[width=0.9\textwidth]{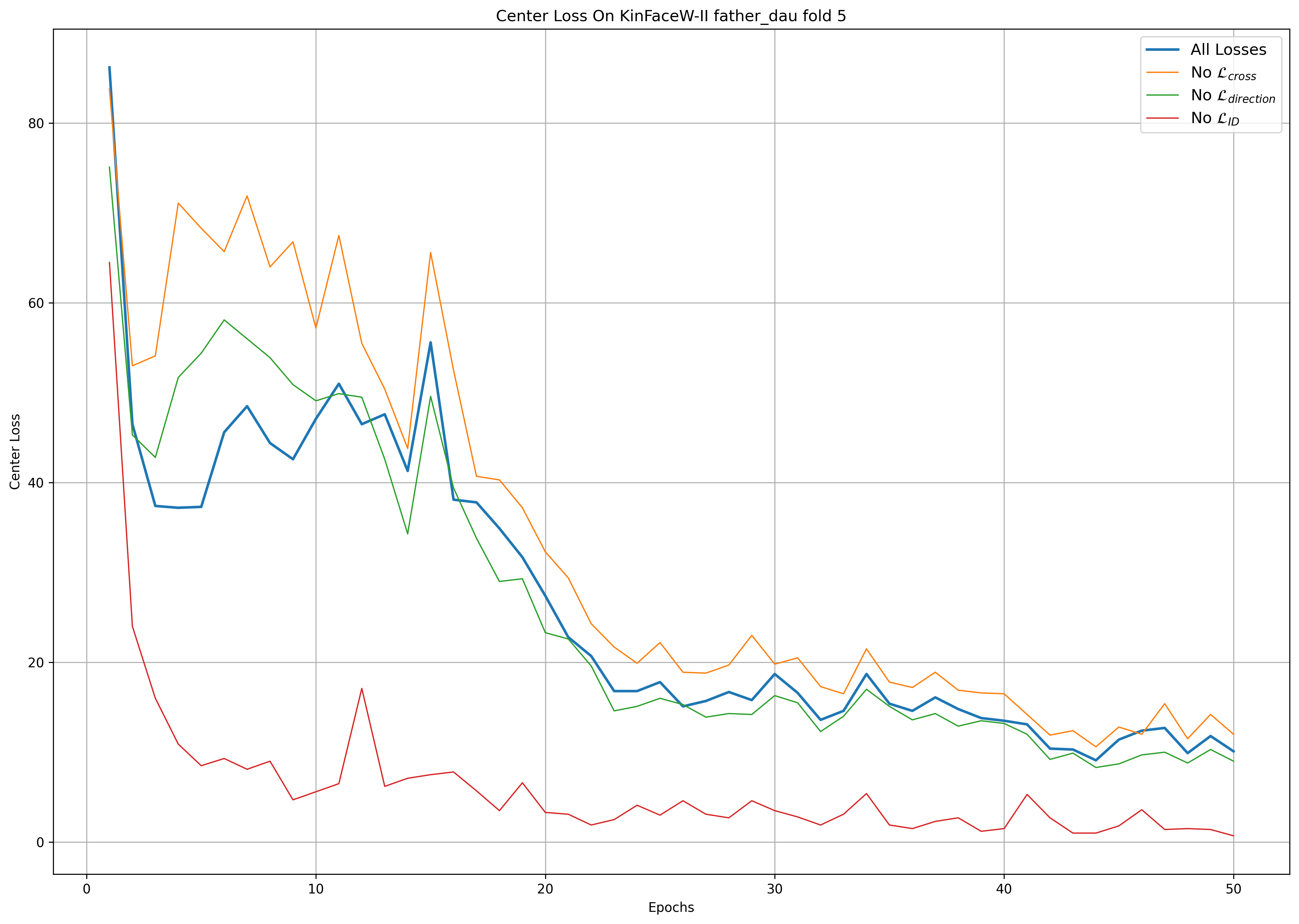}
		\caption{The effect of the center loss for the F-D relation of KinFaceW-II. The blue colour depicts having all losses. The orange, green, and red colours show all losses without $\mathcal{L}_{cross}$, $\mathcal{L}_{direction}$, and $\mathcal{L}_{ID}$, respectively. The figures from the top-left corner belong to the test folds 1, 2, 3, 4, and 5, in order.
		}
		\label{fig:visualization-center-loss}
	\end{center}
\end{figure}

The effect of the center loss for the F-D relation of KinFaceW-II for the five test folds was depicted in Fig.~\ref{fig:visualization-center-loss}. All losses and the absence of $\mathcal{L}_{cross}$, $\mathcal{L}_{direction}$, and $\mathcal{L}_{ID}$, were investigated.

\subsection{The test accuracy figures of the five folds}
\label{appendix:performance-visualization-test-accuracy}

\begin{figure}[pos=htbp]
	\begin{center}
		\includegraphics[width=0.45\textwidth]{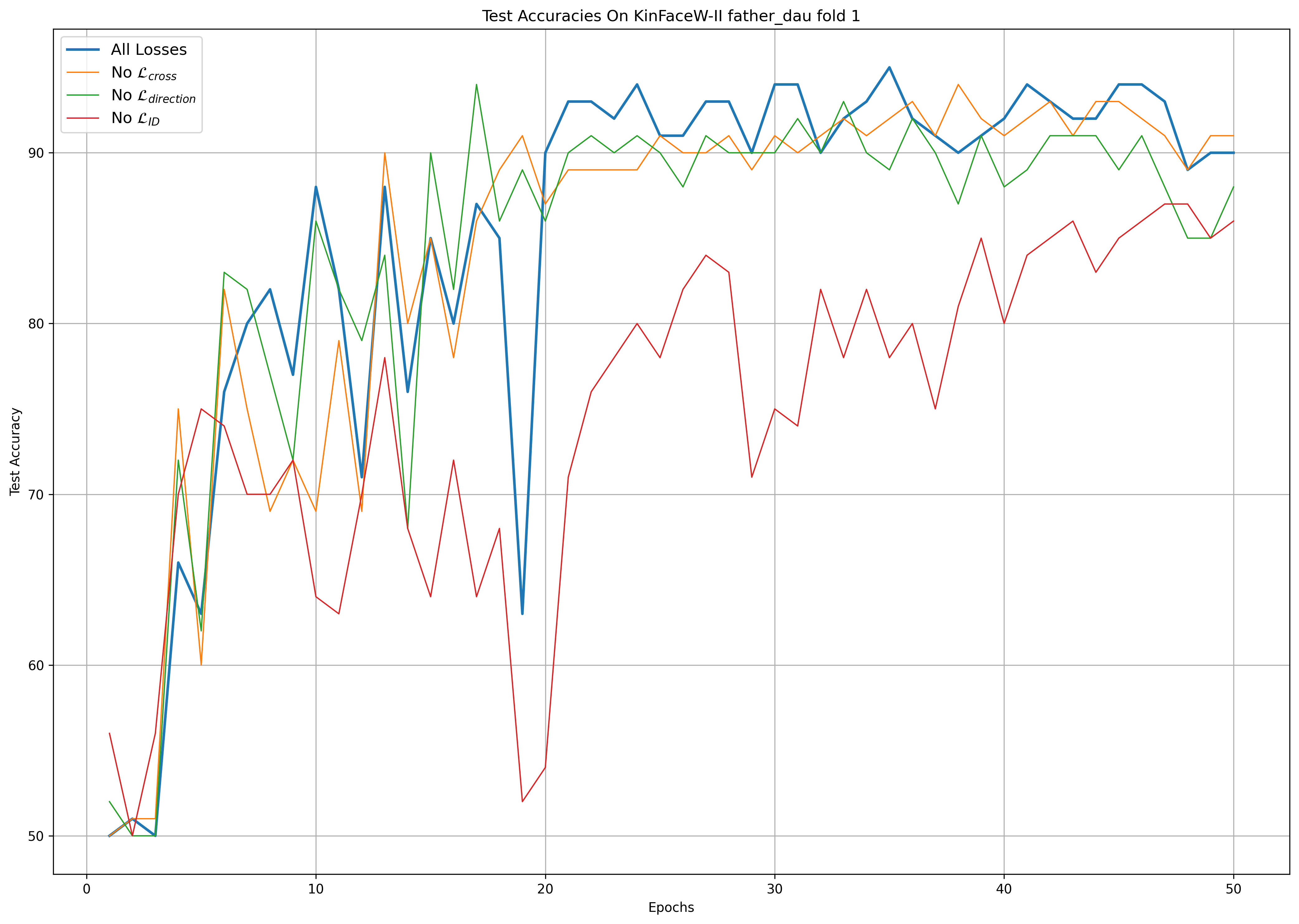}
		\hspace{0.05\textwidth} 
		\includegraphics[width=0.45\textwidth]{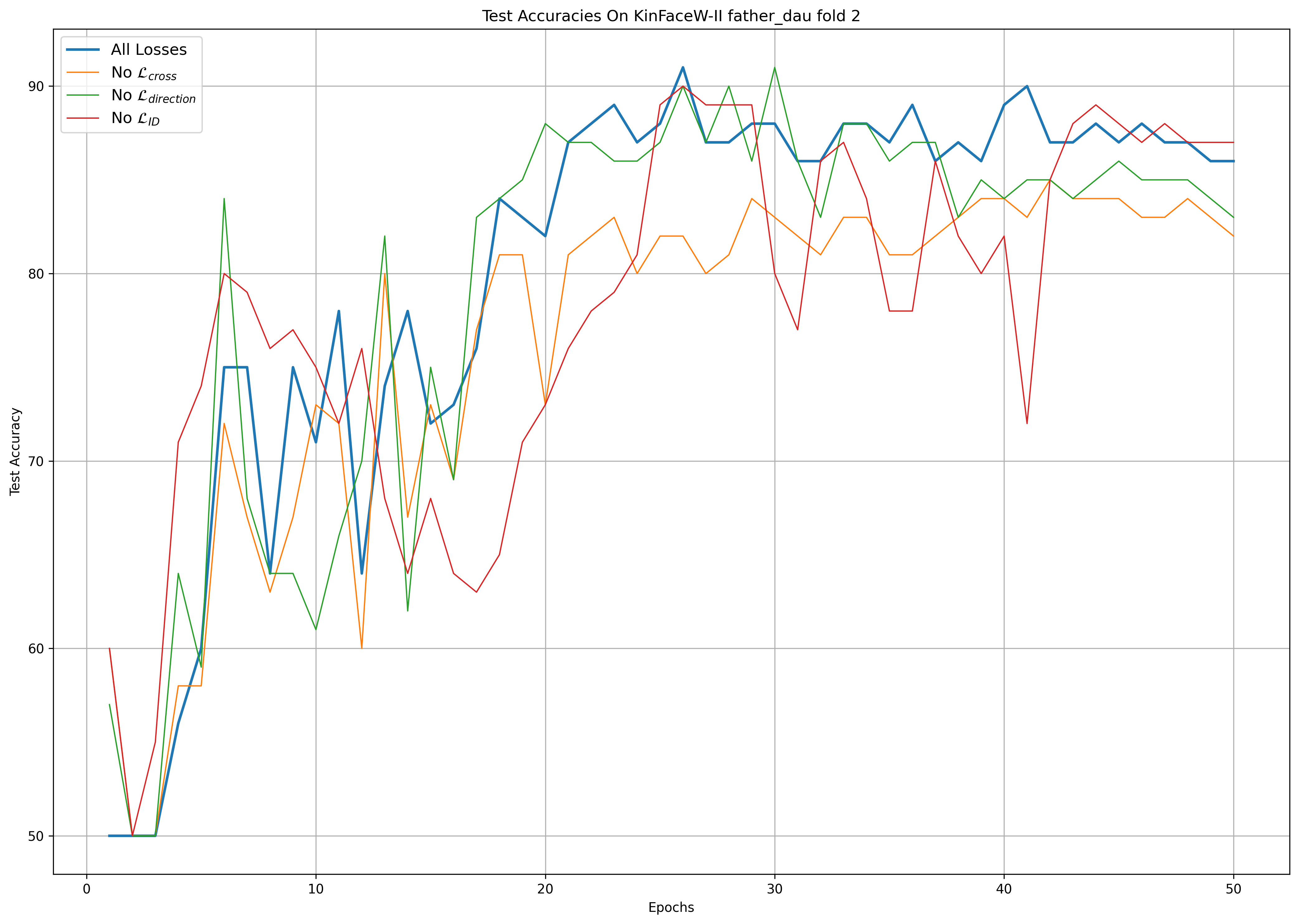}
		\hspace{0.05\textwidth} 
		\includegraphics[width=0.45\textwidth]{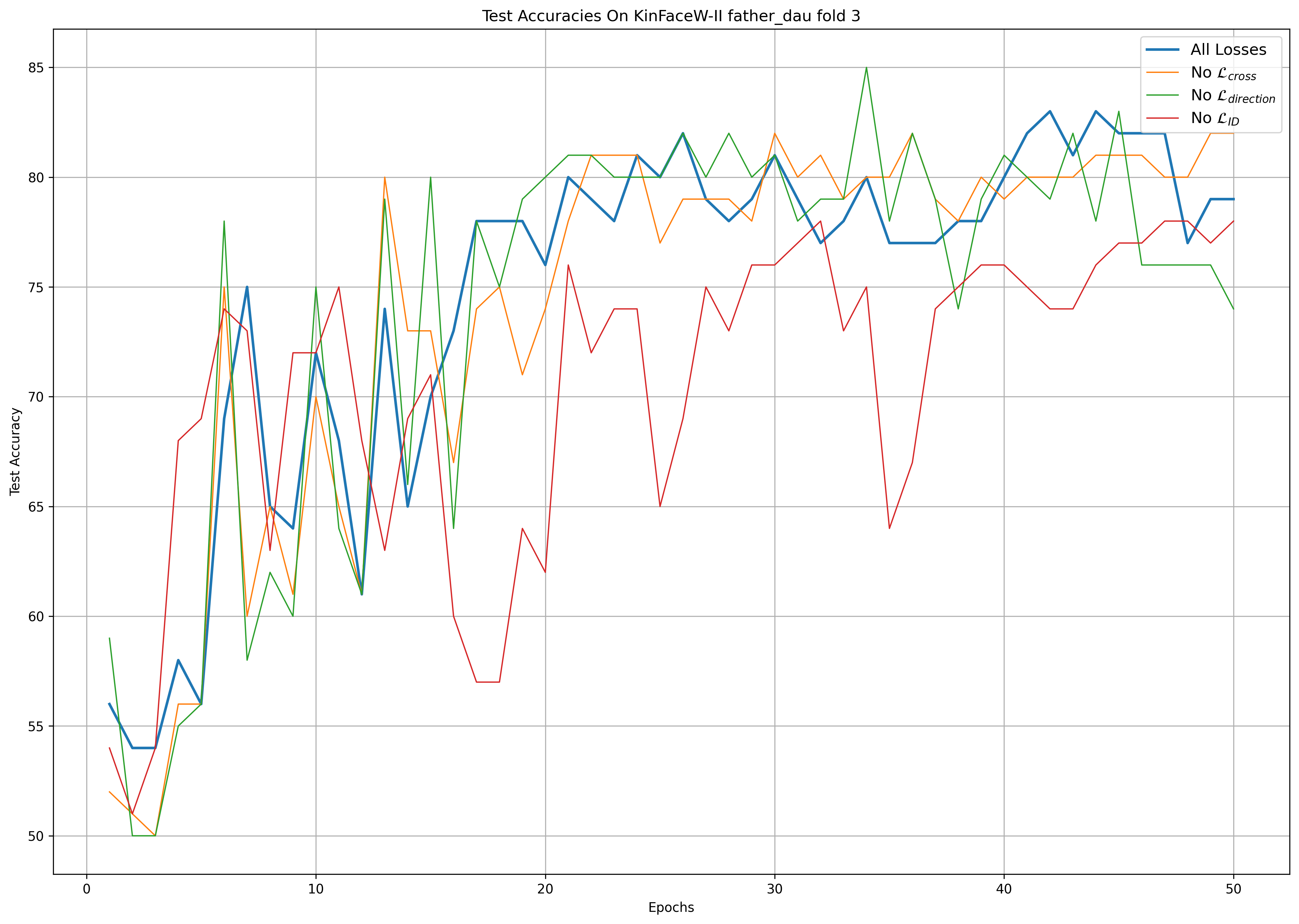}
		\hspace{0.05\textwidth} 
		\includegraphics[width=0.45\textwidth]{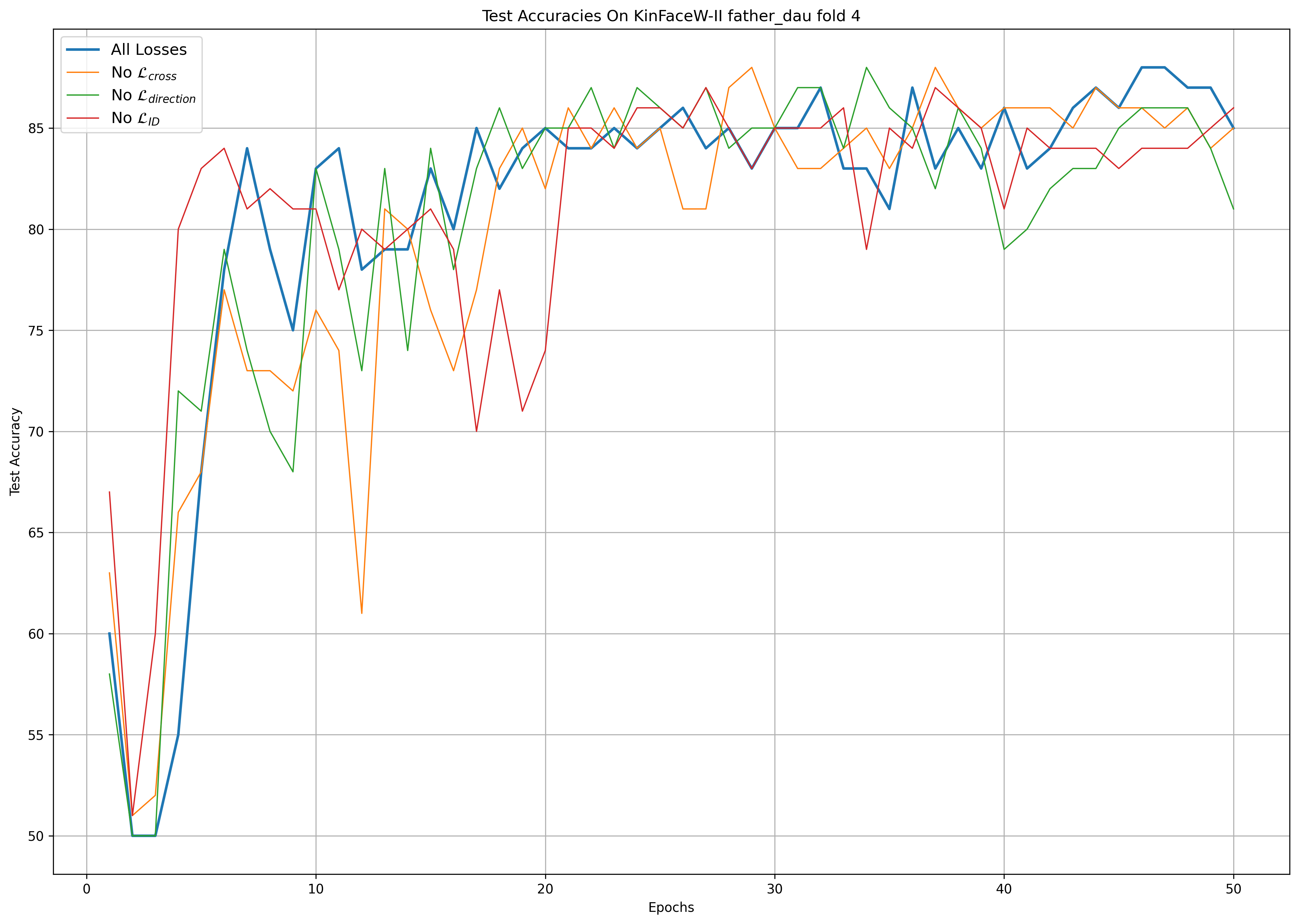}
		\hspace{0.05\textwidth} 
		\includegraphics[width=0.90\textwidth]{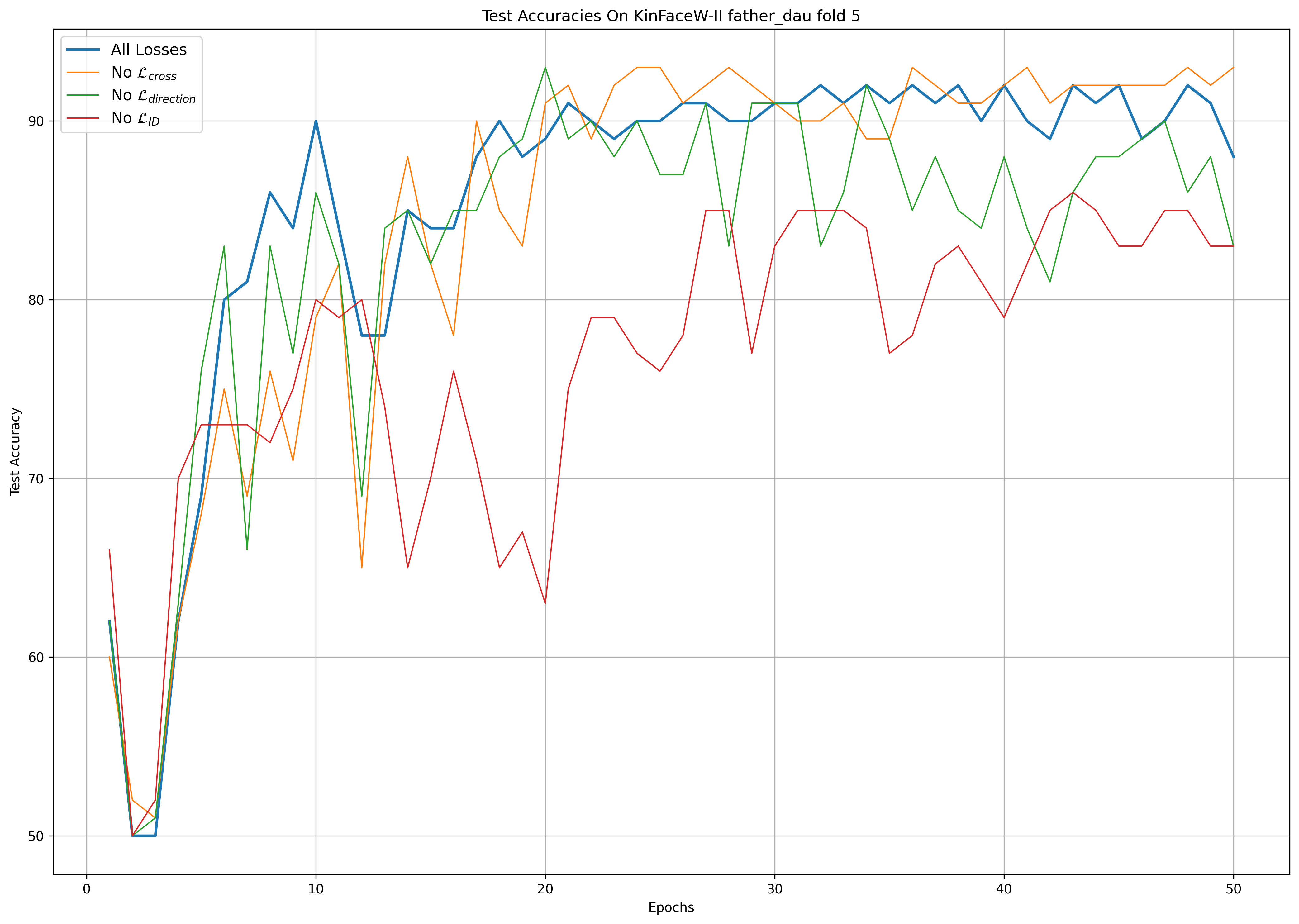}
		\caption{Test accuracy figures for the five folds of the F-D relation on KinFaceW-II. The blue colour depicts having all losses. The orange, green, and red colours show all losses without $\mathcal{L}_{cross}$, $\mathcal{L}_{direction}$, and $\mathcal{L}_{ID}$, respectively. The figures from the top-left corner belong to the test folds 1, 2, 3, 4, and 5, in order.
		}
		\label{fig:visualization-test-accuracy}
	\end{center}
\end{figure}

The effect of the center loss in elavationg the KV accuracy of the F-D relation of KinFaceW-II was depicted in Fig.~\ref{fig:visualization-test-accuracy}.

\section{Effect of Each Facial Component and its Occlusion}
\label{appendix:effect-each-component}%

We reported the KV accuracy on KinFaceW-II for the presence or absence of facial components in the graph in Table~\ref{table:effect-each-component}. The first row presented the presence of all nodes. Based on our definition in Fig.~\ref{fig:main_components} for occluded and facial component images, rows 2-6 showed the results when information of one facial component in the graph was missing. Specifically, this means that the embedding of one of the four facial component images for both the parent and the child was missing. 

Rows 7-10 in Table~\ref{table:effect-each-component} illustrate the results when the information of an occluded node was missing. In this context, an occluded node referred to the embedding of one of four occluded images of the parent and child, as shown in the top row of Fig.~\ref{fig:main_components}. Overall, there were four occluded nodes for each of the parent and child, each representing an image without a facial component.

Row 11 presented a graph of ten nodes: five nodes related to the parent landmarks and five nodes to the child: one for the entire face and four face images with one facial component occluded. Row 12 showed a graph of ten nodes. Each person contained one for the entire face and four facial components.

Row 13 represented the effect of information from the entire face. The graph had only two nodes.
Row 14 reported when information about a person's left and right eyes was preserved. Its forest graph had four nodes.
Rows 15-16 represented the effect of nose and mouth components.
Rows 17-19 illustrated the impact of occlusion.

\begin{table}[pos=htbp]
	\caption{Effect of each facial component and missing one component on KinFaceW-II.}
	\begin{tabular}{l|l|llll|l}
		& Region                                         & F-S  & F-D  & M-D  & M-S  & Mean  \\ \toprule
		1  & All nodes                                      & 94.2 & 90   & 96.2 & 92.6 & 93.25 \\
		2  & All nodes without the entire face              & 94.2 & 90.4 & 96.2 & 92.6 & 93.35 \\
		3  & All nodes without the left eye                 & 94.2 & 90.2 & 95.8 & 92.8 & 93.25 \\
		4  & All nodes without the right eye                & 94.0 & 90.0 & 96.0 & 92.0 & 93    \\
		5  & All nodes without the nose                     & 94.2 & 89.8 & 96.2 & 92.6 & 93.2  \\
		6  & All nodes without the mouth                    & 93.6 & 89.2 & 96.0 & 92.8 & 92.9  \\
		7  & All nodes without the left-eye occluded image  & 94.4 & 89.8 & 96.2 & 92.4 & 93.2  \\
		8  & All nodes without the right-eye occluded image & 93.8 & 89.2 & 96.2 & 92.6 & 92.95 \\
		9  & All nodes without the nose occluded image      & 94.2 & 90.2 & 96.2 & 92.8 & 93.35 \\
		10 & All nodes without the mouth occluded image     & 94.6 & 90.0 & 95.6 & 92.2 & 93.1  \\
		11 & entire + all four occluded images              & 93.0 & 87.6 & 94.6 & 90.8 & 91.5  \\
		12 & entire + all four facial component images      & 93.8 & 89.6 & 96.0 & 92.4 & 92.95 \\
		13 & only entire                                    & 89.0 & 80.8 & 92.2 & 87.6 & 87.4  \\
		14 & only the left and right eyes                       & 87.8 & 85.0 & 92.8 & 89.2 & 88.7  \\
		15 & only nose                                      & 90.8 & 84.6 & 94.0 & 89.8 & 89.8  \\
		16 & only mouth                                     & 85.2 & 78.4 & 88.4 & 84.2 & 84.05 \\
		17 & only occluded eyes                             & 92.6 & 87.8 & 93.4 & 91.2 & 91.25 \\
		18 & only occluded nose                             & 77.6 & 75.4 & 79.2 & 76.8 & 77.25 \\
		19 & only occluded mouth                            & 68.4 & 68.4 & 64.2 & 60.8 & 65.45
	\end{tabular}
	\label{table:effect-each-component}
\end{table}

\printcredits

\section*{Declaration of competing interest}
The authors declare that they have no known competing financial interests or personal relationships that could have appeared to influence the work reported in this paper. This article has also never been submitted to more than one journal for simultaneous consideration. This article is original.

\section*{Consent for publication}
All authors have read and approved the final manuscript for publication.

\section*{Data availability}
The datasets used in this paper are publicly available.

\section*{Code availability}
The code is available at~\href{https://github.com/ali-nazari/Kinship-Verification}{Github} 

\bibliographystyle{cas-model2-names}

\bibliography{cas-refs}
%
%
%
%
%

\end{document}